\documentclass[10pt,twocolumn,letterpaper]{article}

\usepackage[pagenumbers]{cvpr} 

\usepackage{graphicx}
\usepackage{amsmath}
\usepackage{amssymb}
\usepackage{booktabs}

\usepackage[pagebackref,breaklinks,colorlinks]{hyperref}
\makeatletter
\newcommand{\printfnsymbol}[1]{%
  \textsuperscript{\@fnsymbol{#1}}%
}
\makeatother
\usepackage{makecell}
\usepackage{multirow}
\usepackage{caption}
\usepackage[capitalize]{cleveref}
\crefname{section}{Sec.}{Secs.}
\Crefname{section}{Section}{Sections}
\Crefname{table}{Table}{Tables}
\crefname{table}{Tab.}{Tabs.}

\def\cvprPaperID{6022} 
\def\confName{CVPR}
\def\confYear{2022}

\newcommand{\class}[1]{\ensuremath{\mathsf{#1}}}
\newcommand{\std}[1]{\scriptsize{$\pm$#1}}
\newcommand{\xhdr}[1]{\vspace{2mm}\noindent{{\bf #1.}}}

\begin{document}

\title{Estimating Example Difficulty using Variance of Gradients}

\author{Chirag Agarwal\thanks{Equal Contribution}\\
MDSR Lab, Adobe\\
{\tt\small chiragagarwall12@gmail.com}
\and
Daniel D'souza\printfnsymbol{1}\\
  ML Collective\\
{\tt\small ddsouza@umich.edu}
\and
 Sara Hooker \\
   Google Research\\
{\tt\small shooker@google.com}
}
\maketitle

\begin{abstract}
    \looseness=-1
   In machine learning, a question of great interest is understanding what examples are challenging for a model to classify. Identifying atypical examples ensures the safe deployment of models, isolates samples that require further human inspection and provides interpretability into model behavior. In this work, we propose Variance of Gradients (VoG)\footnote{Code and downloadable VoG scores at \url{https://varianceofgradients.github.io/}. Correspondence to: \textit{Chirag Agarwal}~(chiragagarwall12@gmail.com)} as a valuable and efficient metric to rank data by difficulty and to surface a tractable subset of the most challenging examples for human-in-the-loop auditing. We show that data points with high VoG scores are far more difficult for the model to learn and over-index on corrupted or memorized examples. Further, restricting the evaluation to the test set instances with the lowest VoG improves the model's generalization performance. Finally, we show that VoG is a valuable and efficient ranking for out-of-distribution detection.
\end{abstract}

\section{Introduction}
\label{sec:intro}
Over the past decade, machine learning models are increasingly deployed to high-stake decision applications such as healthcare \cite{badgeley2019deep,Gruetzemacher20183DDL,oakden2020hidden,xie2019automated}, self-driving cars \cite{2017Telsa} and finance \cite{ozbayoglu2020deep}.
For gaining trust from stakeholders and model practitioners, it is important for deep neural networks (DNNs) to make decisions that are interpretable to both researchers and end-users. To this end, for sensitive domains, there is an urgent need for auditing tools which are scalable and help domain experts audit models.

Reasoning about model behavior is often easier when presented with a subset of data points that are relatively more difficult for a model to learn. Besides aiding interpretability through case-based reasoning \cite{caruana2000case,kim2016examples,hooker2019compressed}, it can also be used to surface a tractable subset of atypical examples for further human auditing \cite{leibig2017leveraging,zhang1992selecting}, for active learning to inform model improvements, and to choose not to classify some instances when the model is uncertain \cite{Bartlett2008, cortes2016boosting,GUHAROY2022102274}.
One of the biggest bottlenecks for human auditing is the large scale of modern datasets and the cost of annotating individual features \cite{veale2017fairer,Khan_2021,mckane2021}. Methods which automatically surface a subset of relatively more challenging examples for human inspection help prioritize limited human annotation and auditing time. Despite the urgency of this use-case, ranking examples by difficulty has had limited treatment in the context of deep neural networks due to the computational cost of ranking a high dimensional feature space.

\looseness=-1
\xhdr{Present work} A popular interpretability tool is saliency maps, where each of the features of the input data are scored based on their contribution to the final output \cite{simonyan2014deep}. However, these explanations are typically for a single prediction and generated after the model is trained. Our goal is to leverage these explanations to automatically surface a subset of relatively more challenging examples for human inspection to help prioritize limited human annotation and auditing time. To this end, we propose a ranking method across all examples that instead measures the per-example change in explanations over training. Examples that are difficult for a model to learn will exhibit higher variance in gradient updates throughout training. On the other hand, the backpropagated gradients of the samples that are \emph{relatively easier} will exhibit lower variance because the loss from these examples does not consistently dominate the model training.

We term this class normalized ranking mechanism \textit{Variance of Gradients} (VoG) and demonstrate that VoG is a meaningful way for ranking data by difficulty and surfacing a tractable subset of the most challenging examples for human-in-the-loop auditing across a variety of large-scale datasets. VoG assigns higher scores to test set examples that are more challenging for the model to classify and proves to be an efficient tool for detecting out-of-distribution (OoD) samples. VoG is model and domain-agnostic as all that is required is the backpropagated gradients from the model.
    
\xhdr{Contributions} We demonstrate consistent results across two architectures and three datasets -- Cifar-10, Cifar-100  \cite{krizhevsky2009learning} and ImageNet \cite{russakovsky2015imagenet}. Our contributions can be enumerated as follows:
\begin{enumerate}
\itemsep0em
    \item We present Variance of Gradients (VoG) -- a class-normalized gradient variance score for determining the relative ease of learning data samples within a given class (Sec.~\ref{sec:method}). VoG identifies clusters of images with clearly distinct semantic properties, where images with low VoG scores feature far less cluttered backgrounds and more prototypical vantage points of the object (Fig.~\ref{fig:top_10_acc_vog}). In contrast, images with high VoG scores over-index on images with cluttered backgrounds and atypical vantage points of the object of interest.
    \item VoG effectively surfaces memorized examples, \ie it allocates higher scores to images that require \emph{memorization} (Sec.~\ref{sec:vog_misclassify}).
    Further, VoG aids in understanding the model behavior at different training stages and provides insight into the learning cycle of the model.
    \item We show the reliability of VoG as an OoD detection technique and compare its performance to 9 existing OoD methods, where it outperforms several methods, such as PCA \cite{hawkins1974detection} and KDE \cite{davis2011remarks,parzen1962estimation}. VoG presents an overall improvement of $9.26\%$ in precision compared to all other methods.
\end{enumerate}

\begin{figure*}
	\centering
	\begin{subfigure}{0.45\linewidth}
		\centering
    	\includegraphics[width=0.95\columnwidth]{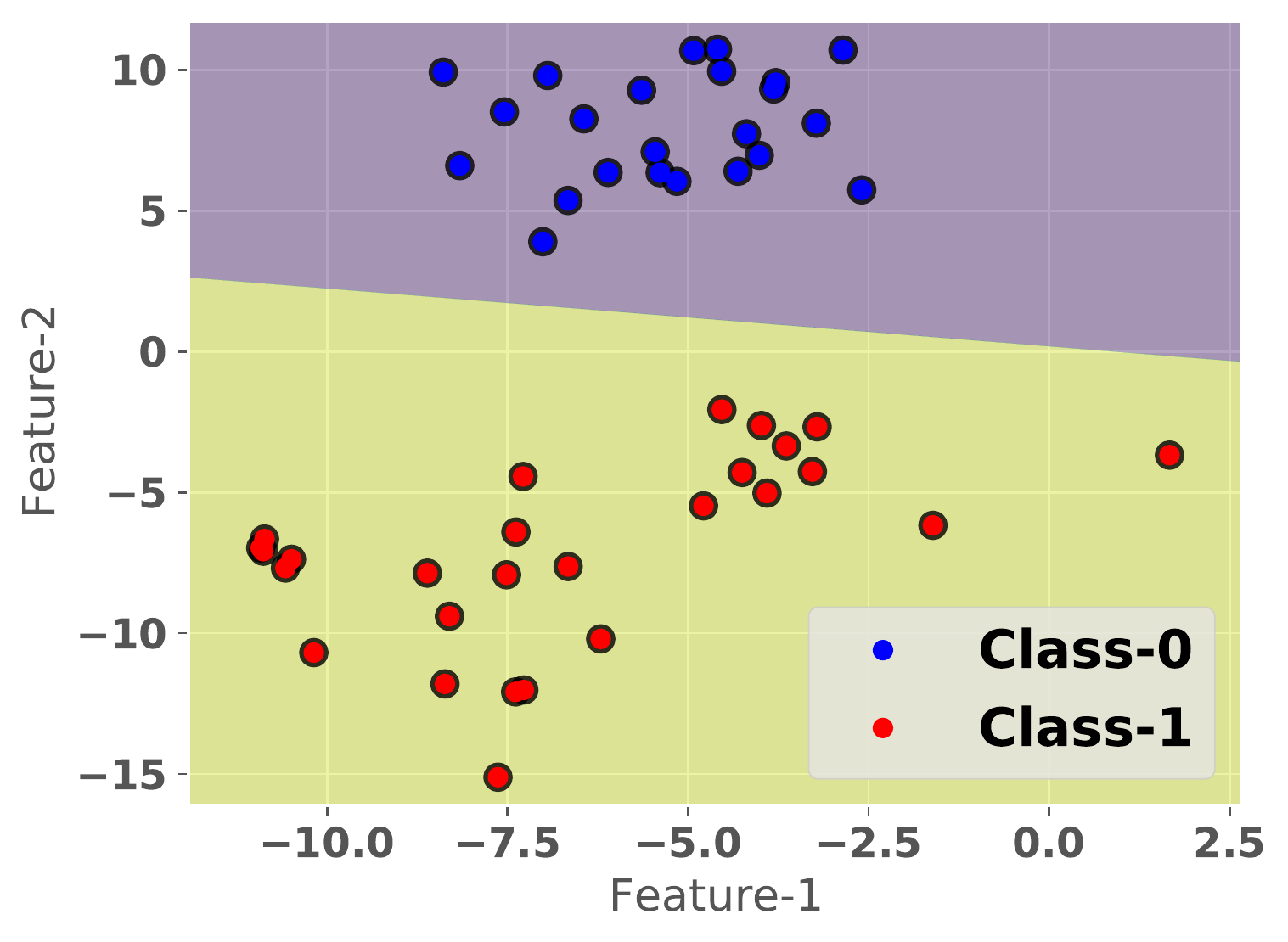}
	    \caption{Toy dataset trained decision boundary}
	    \label{fig:toy_decision_boundary}
	\end{subfigure}
	\begin{subfigure}{0.45\linewidth}
		\centering
    	\includegraphics[width=0.90\columnwidth]{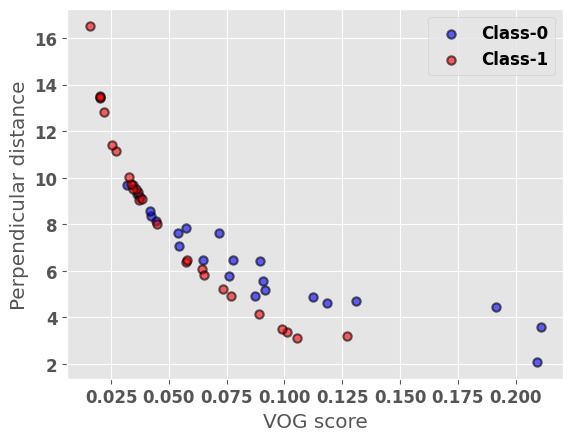}
		\caption{Distance vs. VoG score}
		\label{fig:toy_distance_vog}
	\end{subfigure}
	\caption{
	    \looseness=-1
	    \textbf{Left:} Variance of Gradients (VoG) for each testing data point in the two-dimensional toy problem. \textbf{Right:} VoG accords higher scores to the most challenging examples closest to the decision boundary (as measured by the perpendicular distance). 
	}
	\label{fig:toy_dataset}
\end{figure*}

\begin{figure*}[ht!]
	\centering
            \begin{sc}
   	        {
        	\begin{flushleft}
             \hspace{-0.05cm}\rotatebox{90}{\hspace{-7.7cm}Cifar-10; \class{plane}\hspace{1.6cm}Cifar-100; \class{apple}}
        	    \hspace{0.81cm}Lowest VoG
        	    \hspace{1.8cm}Highest VoG
        	    \hspace{2.4cm}Lowest VoG
        	    \hspace{1.8cm}Highest VoG
        	    \vspace{-0.2cm}
    		\end{flushleft}
    	    }
        	\begin{subfigure}{0.49\linewidth}
        		\centering
          	    \includegraphics[width=0.95\linewidth]{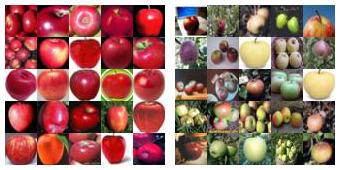}\\
            	\includegraphics[width=0.95\linewidth]{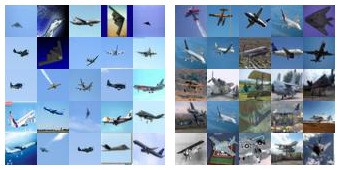}
            	\caption{Early-stage training}
            	\label{fig:color_bias}
        	\end{subfigure}
        	\begin{subfigure}{0.49\linewidth}
        		\centering
            	\includegraphics[width=0.95\linewidth]{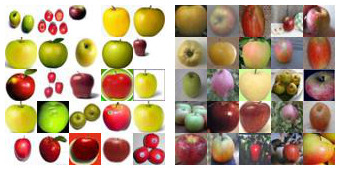}\\
            	\includegraphics[width=0.95\linewidth]{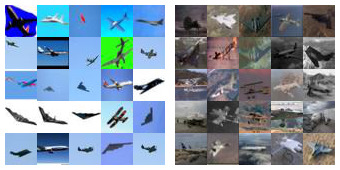}
            	\caption{Late-stage training}
        	\end{subfigure}
        	\caption{
        	    The 5$\times$5 grid shows the top-25 Cifar-10 and Cifar-100 training-set images with the lowest and highest VoG scores in the \emph{Early} (a) and \emph{Late} (b) training stage respectively of two randomly chosen classes. Lower VoG images evidence uncluttered backgrounds (for both \class{apple} and \class{plane}) in the \emph{Late} training stage. VoG also appears to capture a color bias present during the \emph{Early} training stage  for both \class{apple} (red). The VoG images in \emph{Late} training stage present unusual vantage points, with images where the frame is zoomed in on the object of interest.
        	}
        	\label{fig:cifar100_vog_images}
	    \end{sc}
\end{figure*}

\section{VoG Framework}\label{sec:method}
We consider a supervised classification problem where a DNN is trained to approximate the function $\mathcal{F}$ that maps an input variable $\mathbf{X}$ to an output variable $\mathbf{Y}$, formally $\mathcal{F}: \mathbf{X} \mapsto \mathbf{Y}$, where 
$\mathbf{Y}$ is a discrete label vector associated with each input $\mathbf{X}$ and $y\in \mathbf{Y}$ corresponds to one of $C$ categories or classes in the dataset.

A given input image $\mathbf{X}$ can be decomposed into a set of pixels $x_i$, where $i=\{1, \dots, N\}$ and $N$ is the total number of pixels in the image. For a given image, we compute the gradient of the activation $A_{p}^{l}$ with respect to each pixel $x_i$, where $l$ designates the pre-softmax layer of the network and $p$ is the index of either the true or predicted class probability. We would like to note that the pre-softmax layer is responsible for connecting activations from previous layers in the network to individual class scores. Hence, computing the gradients w.r.t. this class indexed score measures the contribution of features to the final class prediction~\cite{simonyan2014deep}.

Note our goal is to rank examples, so for each example, we compute the pre-softmax activation gradient indexed at predicted/true label with respect to the input. This is far more computationally efficient than computing the full Jacobian matrix with individual layers.

Let $\mathbf{S}$ be a matrix that represents the gradient of $A_{p}^{l}$ with respect to individual pixels $x_{i}$, \ie for an image of size $3{\times}32{\times}32$, the gradient matrix $\mathbf{S}$ will be of dimensions $3{\times}32{\times}32$.
\begin{equation}
    \mathbf{S} = \frac{\partial A_p^{l}}{\partial x_i}
\end{equation}
This formulation may feel familiar as it is often computed based upon the weights of a trained model and visualized as a image heatmap for interpretability purposes
\cite{hooker2018benchmark,simonyan2014deep,shrikumar2017learning,smilkov2017smoothgrad,sundararajan2017axiomatic,baehrens2010explain,simonyan2014deep}. In contrast to saliency maps which are inherently local explanation tools, we are leveraging relative changes in gradients across training to rank all examples globally.

Following several seminal papers in explainability literature \cite{hooker2018benchmark,simonyan2014deep,shrikumar2017learning,smilkov2017smoothgrad,sundararajan2017axiomatic}, we take the average over the color channels to arrive at a gradient matrix \cite{shrikumar2017learning,simonyan2014deep,smilkov2017smoothgrad,sundararajan2017axiomatic} where $\mathbf{S} \in \mathbb{R}^{32\times32}$. For a given set of $K$ checkpoints, we generate the above gradient matrix $\mathbf{S}$ for all individual checkpoints, \ie, $\{\mathbf{S}_{1}, \dots, \mathbf{S}_{K}\}$. We then calculate the mean gradient $\mu$ by taking the average of the $K$ gradient matrices. Note, $\mu$ is the mean across different checkpoints and is of the same size as the gradient matrix $\mathbf{S}$. We then calculate the variance of gradients across each pixel as:
\begin{equation}
    \mu = \frac{1}{K}\sum_{t=1}^{K}\mathbf{S}_{t}.
\end{equation}
\begin{equation}
    \text{VoG}_{p} = \sqrt \frac{1}{K}\sum_{t=1}^{K}(\mathbf{S}_{t} -\mu)^{2}.
    \label{eq:vog-p}
\end{equation}
We average the pixel-wise variance of gradients to compute a scalar VoG score for the given input image: 
\begin{equation}
  \text{VoG}=\frac{1}{N}\sum_{t=1}^{N}(\text{VoG}_{p}),
  \label{eq:vog}
\end{equation}
where $N$ is the total number of pixels in a given image. First calculating the pixel-wise variance (Eqn.~\ref{eq:vog-p}) and then average over the pixels (Eqn.~\ref{eq:vog}) is consistent with previous XAI works where the gradients of an input image are computed independently for each pixel in an image \cite{simonyan2014deep,smilkov2017smoothgrad,sundararajan2017axiomatic}.

In order to account for inherent differences in variance between classes, we normalize the absolute VoG score by class-level VoG mean and standard deviation. This amounts to asking: \textit{What is the variance of gradients for a given image with respect to all other exemplars of this class category?}\\


\looseness=-1
\subsection{Validating the behavior of VoG on synthetic data} In Fig.~\ref{fig:toy_decision_boundary}, we illustrate the principle and effectiveness of VoG in a controlled toy example setting. The data was generated using two separate isotropic Gaussian clusters. In such a simple low dimensional problem, the most challenging examples for the model to classify can be quantified by distance to the decision boundary. In Fig.~\ref{fig:toy_decision_boundary}, we visualize the trained decision boundary of a multiple layer perceptron (MLP) with a single hidden layer trained for $15$ epochs. We compute VoG for each training data point and plot final VoG score for each point against the distance to the trained boundary. In Fig.~\ref{fig:toy_distance_vog}, we can see that VoG successfully ranks highest the examples closest to the decision boundary. The most challenging examples exhibit the greatest variance in gradient updates over the course of the training process. In the following sections, we will scale this toy problem and show consistent results across multiple architectures and datasets.

\subsection{Experimental Setup}\label{sec:result}
\xhdr{Datasets} We evaluate our methodology on Cifar-10 and Cifar-100 \cite{krizhevsky2009learning}, and ImageNet \cite{russakovsky2015imagenet} datasets. For all datasets, we compute VoG for both training and test sets.

\xhdr{Cifar Training} We use a ResNet-18 network \cite{he2016deep} for both Cifar-10 and Cifar-100. For each dataset, we train the model for $350$ epochs using stochastic gradient descent (SGD) and compute the input gradients for each sample every $10$ epochs. We implemented standard data augmentation by applying cropping and horizontal flips of input images. We use a base learning rate schedule of $0.1$ and adaptively change to $0.01$ at $150^{\text{th}}$ and $0.001$ at $250^{\text{th}}$ training epochs. The top-1 test set accuracy for Cifar-10 and Cifar-100 were $89.57\%$ and $66.86\%$ respectively.

\xhdr{ImageNet Training} We use a ResNet-50 \cite{he2016deep} model for training on ImageNet. The network was trained with batch normalization \cite{ioffe2015batch}, weight decay, decreasing learning rate schedules, and augmented training data. We train for $32,000$ steps (approximately $90$ epochs) on ImageNet with a batch size of $1024$. We store $32$ checkpoints over the course of training, but in practice observe that VoG ranking is very stable computed with as few as $3$ checkpoints. Our model achieves a top-1 accuracy of $76.68\%$ and top-5 accuracy of $93.29\%$.

\xhdr{Number of checkpoints} The number of checkpoints used to compute VoG balances efficiency for practitioners to use with the robustness of ranking. This can be set by the practitioner, and we note that in practice the last 3 checkpoints are sufficient for a robust VoG ranking (minimal difference when restricting to the last 3 in Figs.~\ref{fig:imagenet_error_plot}b,\ref{fig:cifar100_tse}b,\ref{fig:cifar10_tse}b vs. evaluating on all checkpoints in Fig.~\ref{fig:top_10_acc_vog}). For all experiments, VoG(\emph{early}-stage) is computed using checkpoints from the first 3 epochs and VoG(\emph{late}-stage) is computed using checkpoints from the last 3 epochs. The test set accuracy at the \emph{early}-stage is $44.65\%$, $14.16\%$, and  $51.87\%$ for Cifar-10, Cifar-100, and ImageNet, respectively. In the \emph{late}-stage it is $89.57\%$, $66.86\%$, and $76.68\%$ for Cifar-10, Cifar-100, and ImageNet, respectively.



\section{Utility of VoG as an Auditing Tool}\label{sec:class_importance}

\looseness=-1
In this section, we evaluate the merits of VoG as an auditing tool. Specifically, we (1) present the qualitative properties of images at both ends of the VoG spectrum, (2) measure how discriminative VoG is at separating easy examples from difficult, (3) quantify the stability of the VoG ranking, (4) use VoG as an auditing tool for test dataset, and (5) leverage VoG to understand the training dynamics of a DNN.

\begin{figure*}[h]
	\centering
\begin{sc}
  	{
    	\begin{flushleft}
        	    \hspace{1.2cm}Lowest VoG
        	    \hspace{1.8cm}Highest VoG
        	    \hspace{2.3cm}Lowest VoG
        	    \hspace{1.8cm}Highest VoG   
		\end{flushleft}
		\vspace{-0.2cm}
	}
	\begin{subfigure}{0.49\linewidth}
		\centering
    	\includegraphics[width=0.99\linewidth]{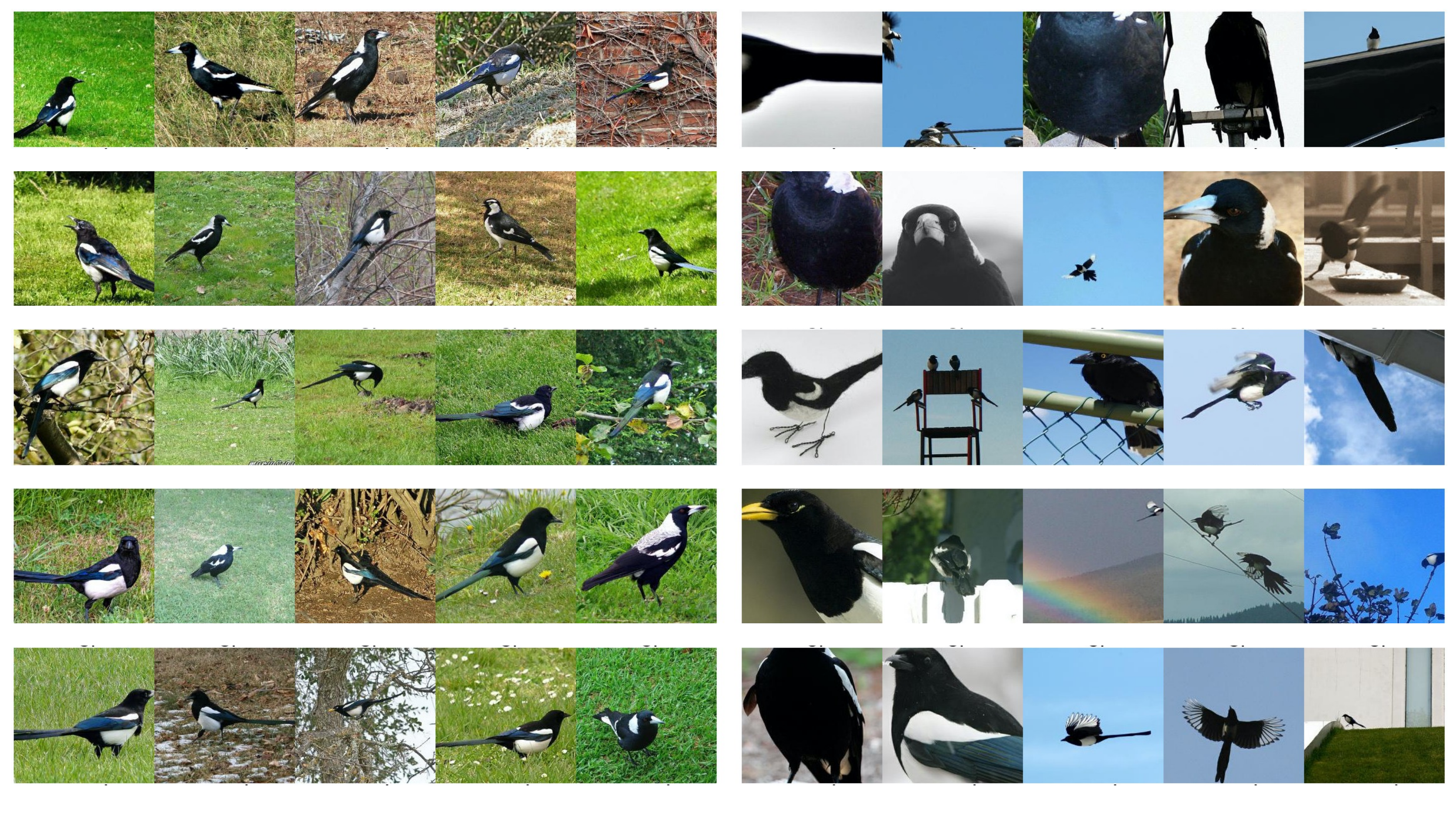}
	\end{subfigure}
	\begin{subfigure}{0.49\linewidth}
		\centering
    	\includegraphics[width=0.99\linewidth]{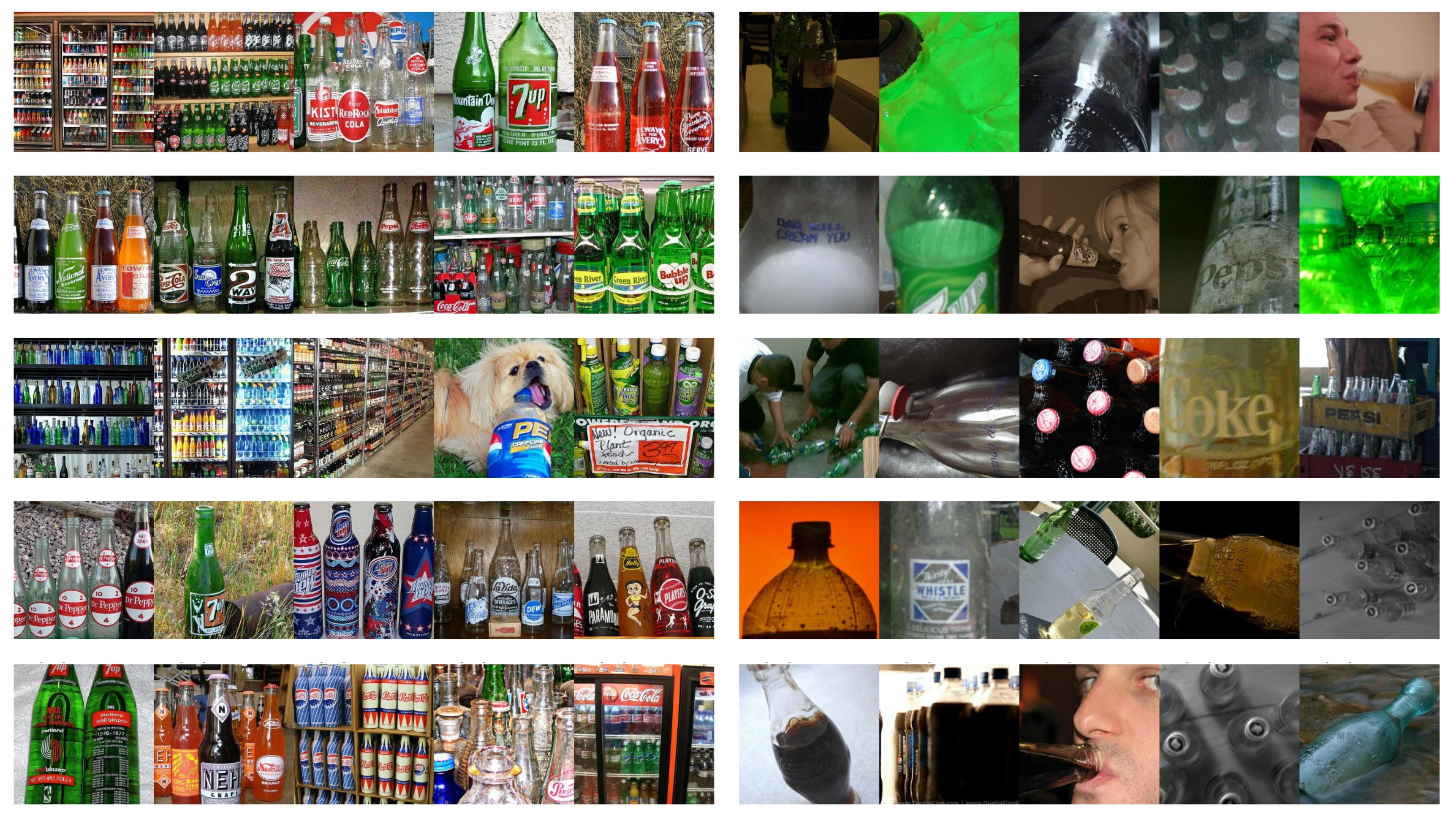}
	\end{subfigure}
	\vspace{-0.2cm}
  	{
    	\begin{flushleft}
    	    \hspace{3.6cm}\texttt{magpie}
    	    \hspace{6.6cm}\texttt{pop~bottle}
		\end{flushleft}
	}
	\end{sc}
	\caption{
	    Each 5$\times$5 grid shows the top-25 ImageNet training-set images with the lowest and highest VoG scores for the class \texttt{magpie} and \texttt{pop bottle}. Training set images with higher VoG scores tend to feature zoomed-in images with atypical color schemes and vantage points.
	}
	\label{fig:imagenet_vog_pop_bottle}
\end{figure*}

\begin{figure*}[h]
    \centering
        \begin{sc}
        	\begin{subfigure}{0.49\linewidth}
        		\centering
            	\includegraphics[width=0.93\linewidth]{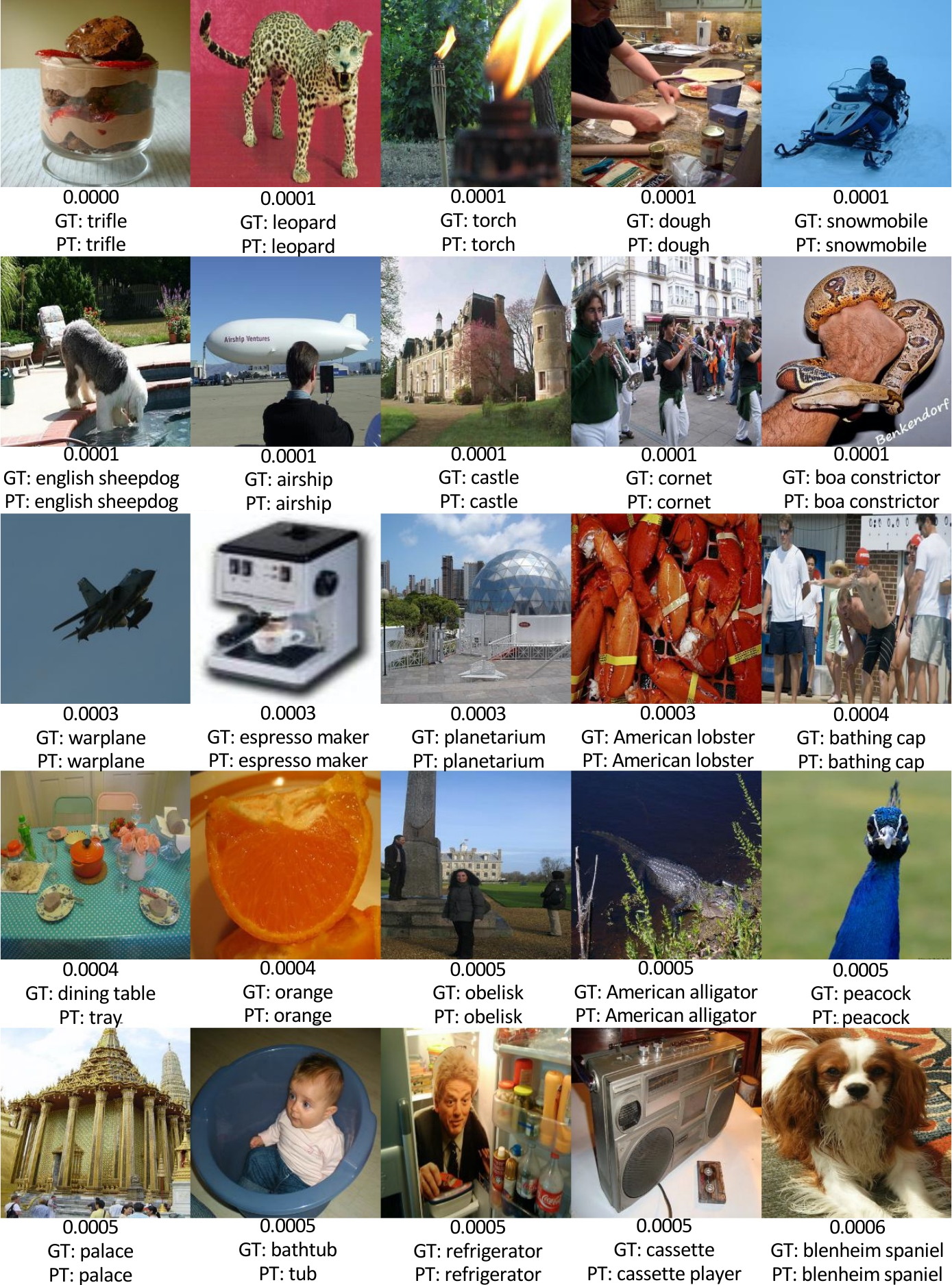}
                \label{fig:vog_predicted_low}
            	\caption{Lowest VoG}        
        	\end{subfigure}
        	\begin{subfigure}{0.49\linewidth}
        		\centering
            	\includegraphics[width=0.93\linewidth]{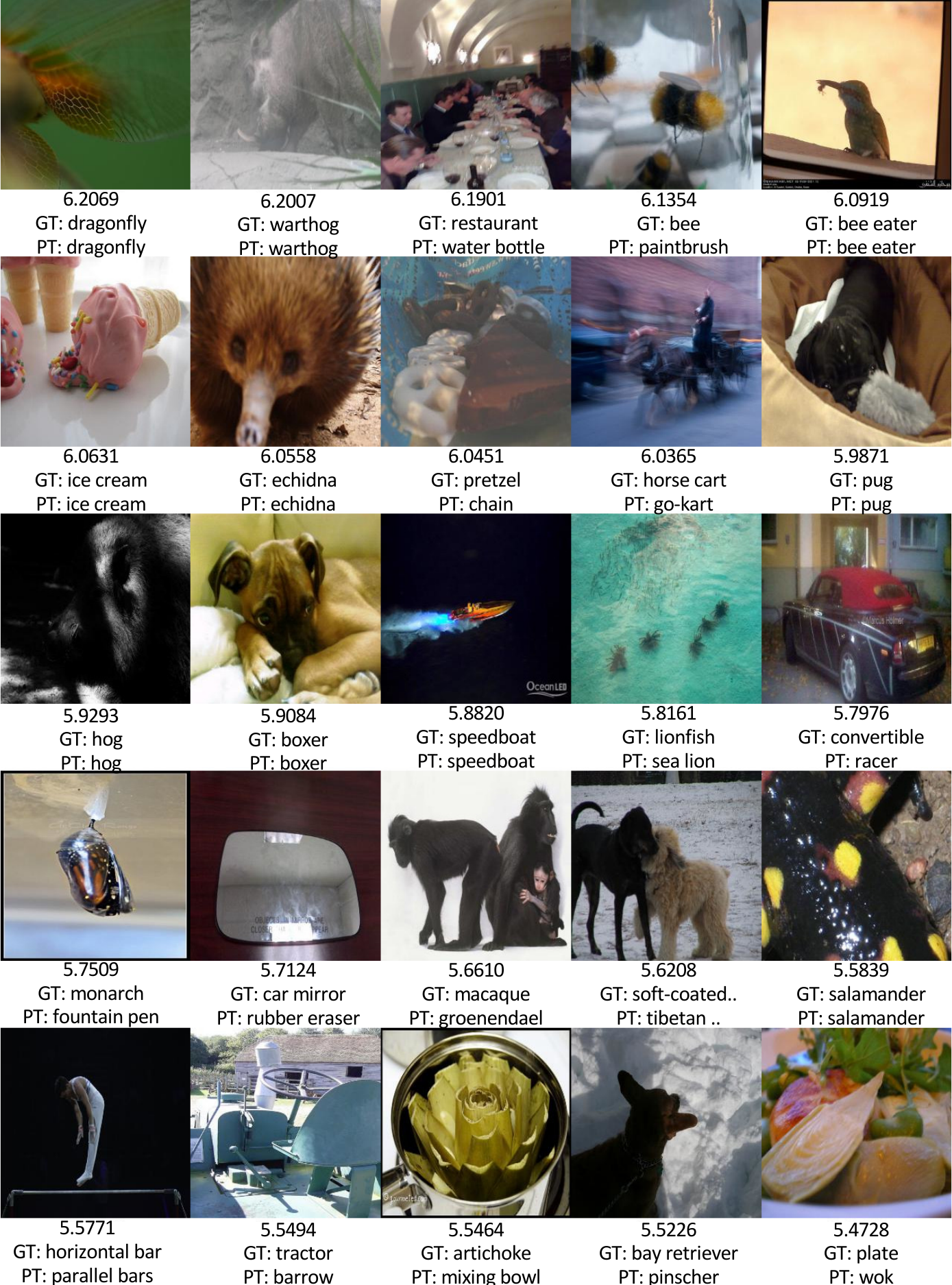}
                \label{fig:vog_predicted_high}
                \caption{Highest VoG}
        	\end{subfigure}
        \end{sc}
    \vskip -0.1in
	\caption{
	    Each 5$\times$5 grid shows the top-25 ImageNet test set images with the lowest and highest VoG scores for the top-1 predicted class. Test set images with higher VoG scores tend to feature zoomed-in images and are misclassified more as compared to the lower VoG images which tend to feature more prototypical vantage points of objects.
	}
	\label{fig:vog_predicted}
\end{figure*}

\xhdr{1) Qualitative inspection of ranking} A qualitative inspection of examples with high and low VoG scores shows that there are distinct semantic properties to the images at either end of the ranking. We visualize $25$ images ranked lowest and highest according to VoG for both the entire dataset (visualized for ImageNet in Fig.~\ref{fig:vog_predicted}) and for specific classes (visualized for ImageNet in Fig.~\ref{fig:imagenet_vog_pop_bottle} and for Cifar-10 and Cifar-100 in Fig.~\ref{fig:cifar100_vog_images}). Images with \emph{low} VoG score tend to have uncluttered and often white backgrounds with the object of interest centered clearly in the frame. Images with the \emph{high} VoG scores have cluttered backgrounds and the object of interest is not easily distinguishable from the background. We also note that images with high VoG scores tend to feature atypical vantage points of the objects such as highly zoomed frames, side profiles of the object or shots taken from above. Often, the object of interest is partially occluded or there are image corruptions present such as heavy blur.

\begin{figure*}[h]
	\centering
	\begin{subfigure}{0.32\linewidth}
		\centering
            \begin{sc}
    	        \includegraphics[width=0.95\linewidth]{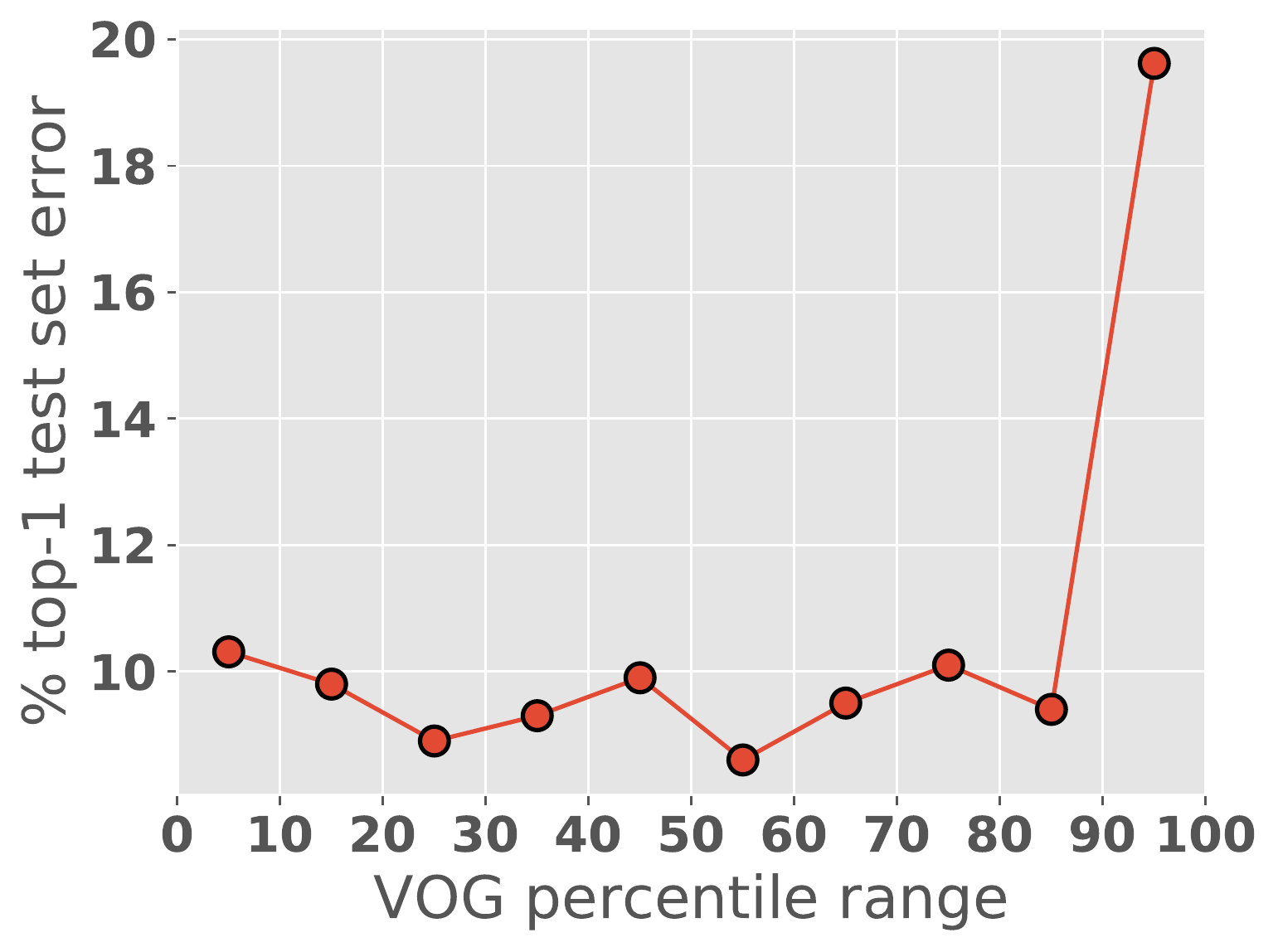}
                \label{fig:top_10_acc_vog_cifar10}
		  	    \caption{Cifar-10}                
            \end{sc}
	\end{subfigure}
	\begin{subfigure}{0.32\linewidth}
		\centering
            \begin{sc}
    	        \includegraphics[width=0.95\linewidth]{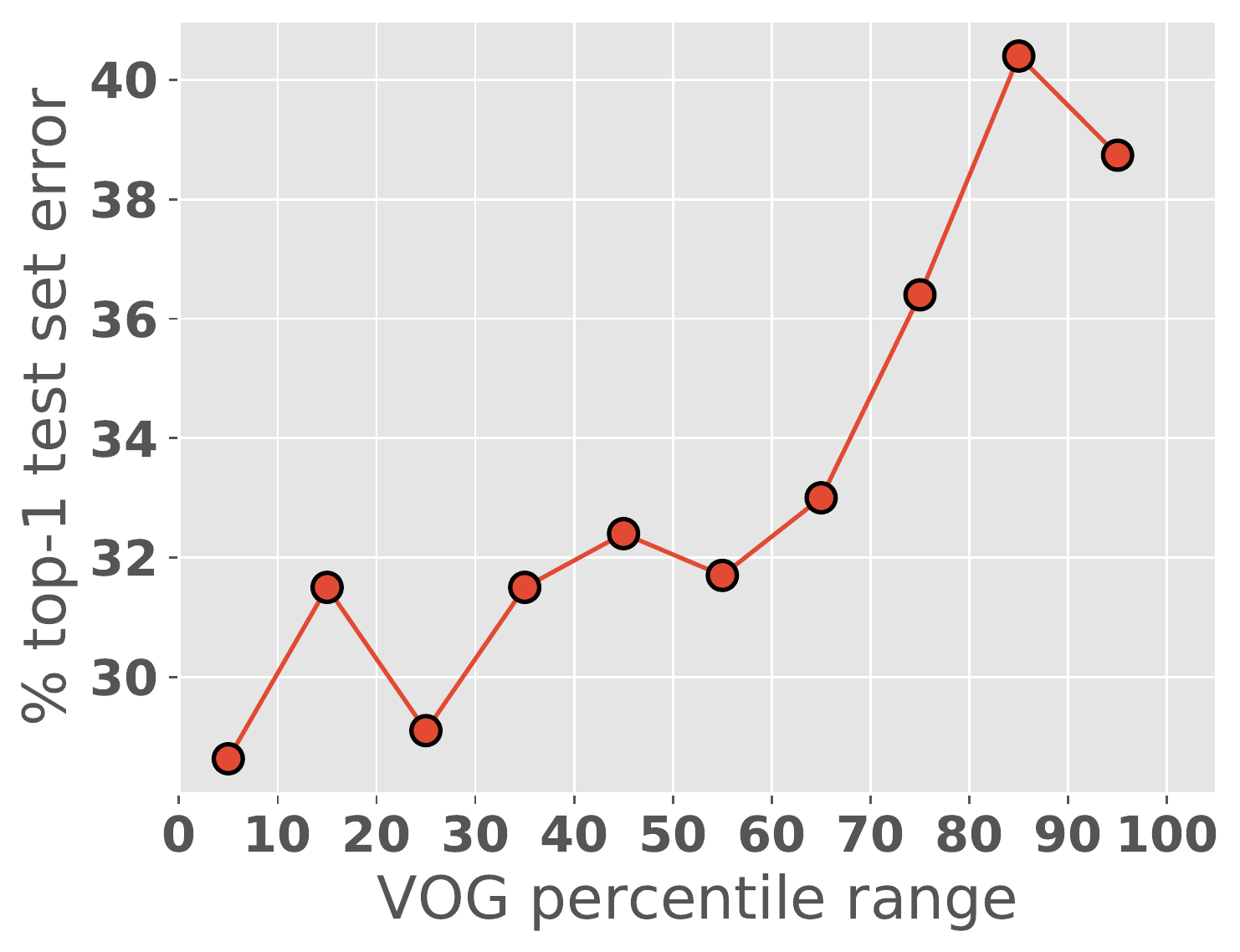}
                \label{fig:top_10_acc_vog_cifar100}
		 	    \caption{Cifar-100}                
             \end{sc}
	\end{subfigure}
	\begin{subfigure}{0.33\linewidth} 
		\centering
            \begin{sc}
    	        \includegraphics[width=0.95\linewidth]{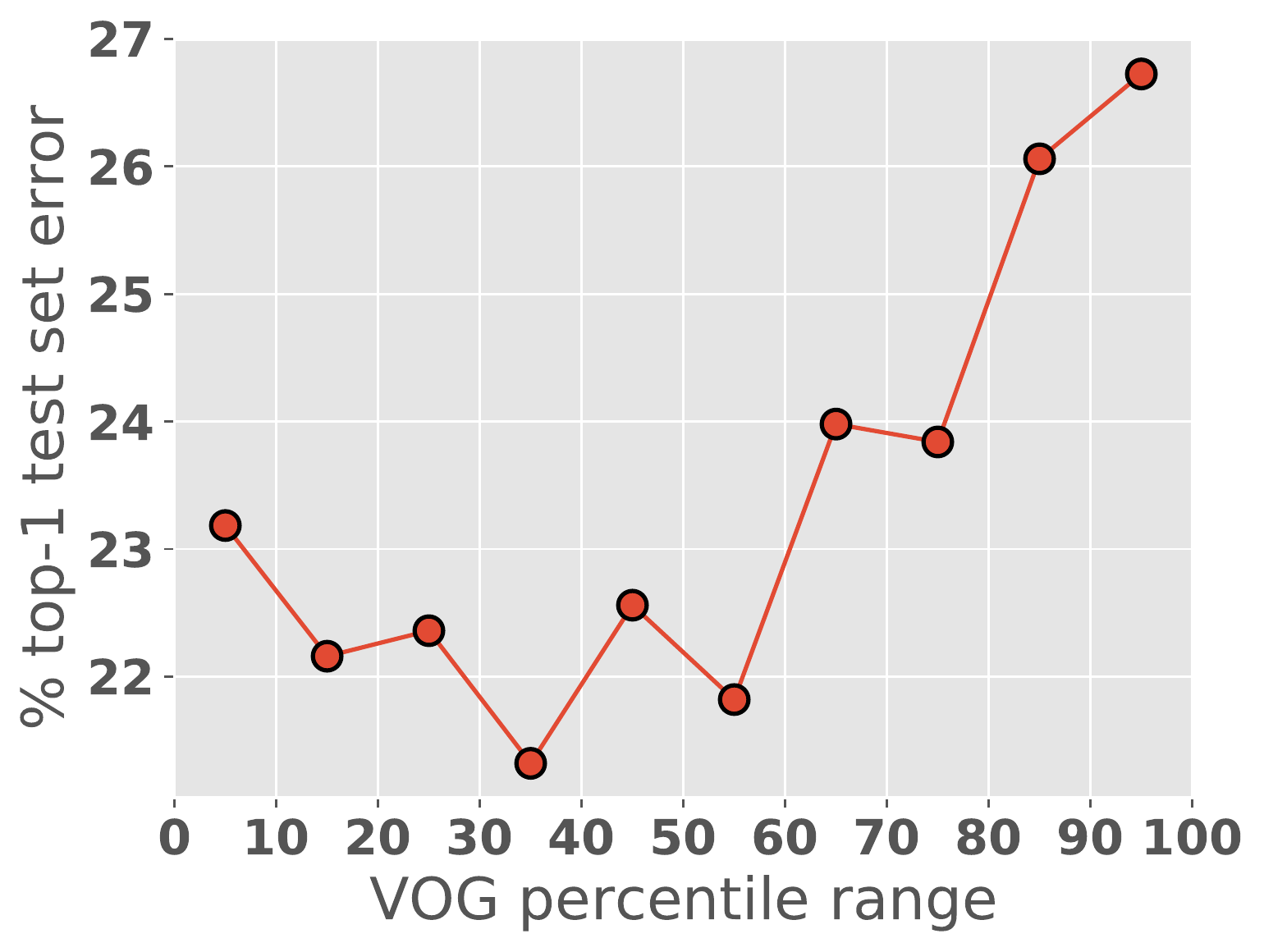}
    	        \label{fig:top_10_acc_vog_imagenet}
		        \caption{ImageNet}    	        
    	    \end{sc}
	\end{subfigure}
	\caption{
	    The mean top-1 test set error (y-axis) for the examples thresholded by VoG score percentile (x-axis). Across Cifar-10, Cifar-100 and ImageNet, mis-classification increases with an increase in VoG scores. Across all datasets the group of samples in the top-10 percentile VoG scores have the highest error rate, \ie contains most number of misclassified samples.
	}
	\label{fig:top_10_acc_vog}
\end{figure*}

\xhdr{2) Test set error and VoG} A valuable property of an auditing tool is to effectively discriminate between easy and challenging examples. 
In Fig.~\ref{fig:top_10_acc_vog}, we plot the test set error of examples bucketed by VoG decile. Note that we plot error, so lower is better. We show that examples at the lowest percentiles of VoG have low error rates, and misclassification increases with an increase in VoG scores. Our results are consistent across all datasets, yet the trend is more pronounced for more complex datasets such as Cifar-100 and ImageNet. We ascribe this to differences in underlying model complexity. Furthermore, in  Fig.~\ref{fig:bar_error_rates}, we observe that test set error on the lowest VoG scored images are lower than the baseline test set performance. 

\begin{figure}[h]
\centering
\begin{sc}
	\centering
	\includegraphics[width=0.81\linewidth]{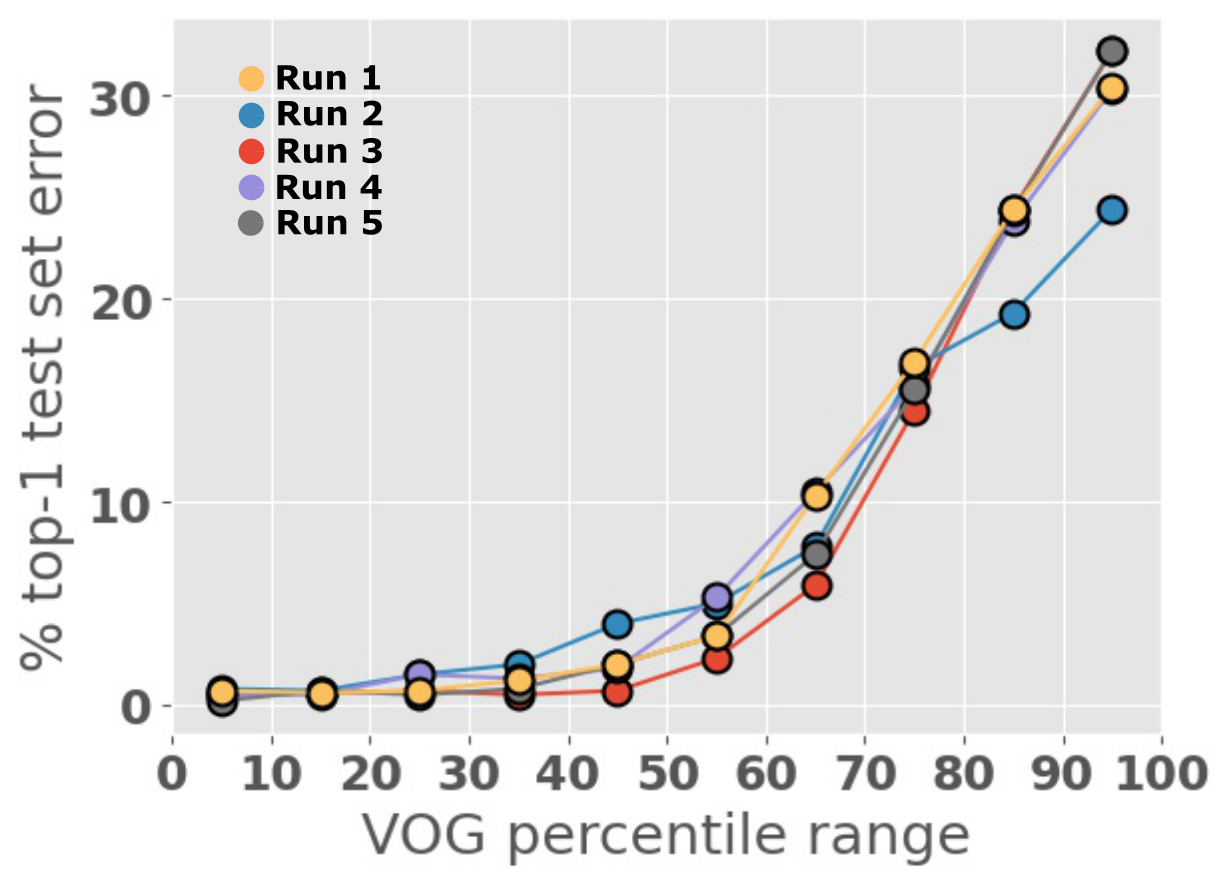}
    \vskip -0.1in
	\caption{
	    The VoG top-1 test set error for five ResNet-18 networks independently trained on Cifar-10 from random initialization. The plot shows that VoG produces a stable ranking with a similar distribution of error in each percentile across all images
	}
	\label{fig:stability_normalized}
	\end{sc}
\end{figure}

\xhdr{3) Stability of VoG ranking} To build trust with an end-user, a key desirable property of any auditing tool is consistency in performance. We would expect a consistent method to produce a ranking with a closely bounded distribution of scores across independently trained runs for a given model and dataset. To measure the consistency of the VoG ranking, we train five Cifar-10 networks from random initialization following the training methodology described in Sec.~\ref{sec:result}. Empirically, Fig.~\ref{fig:stability_normalized} shows that VoG rankings evidence a consistent distribution of test-error at each percentile given the same model and dataset. For completeness, we also measure instance-wise VoG stability by computing the standard deviation of VoG scores for 50k Cifar-10 samples across 10 independent initializations. The standard deviation of the VoG scores is negligible with a mean deviation of ${3.81e^{-9}}$ across all samples. In addition, we find similar results for Cifar-100 dataset where the output VoG scores are stable (mean std of $9.6e{-}6$) across different model initializations. Finally, we extend our stability experiments to understand the effect of different training hyperparameter settings (e.g., batch size) on the VoG scores. Here, we train 5 Cifar-10 models using different batch sizes, i.e., \{128, 256, 384, 512, 640\}, and find that the mean VoG standard deviation across 50k Cifar-10 samples was $1.9e{-}5$.

\begin{figure*}[h]
\centering
\begin{sc}
	\begin{subfigure}{0.49\linewidth}
		\centering
    	\includegraphics[width=0.81\linewidth]{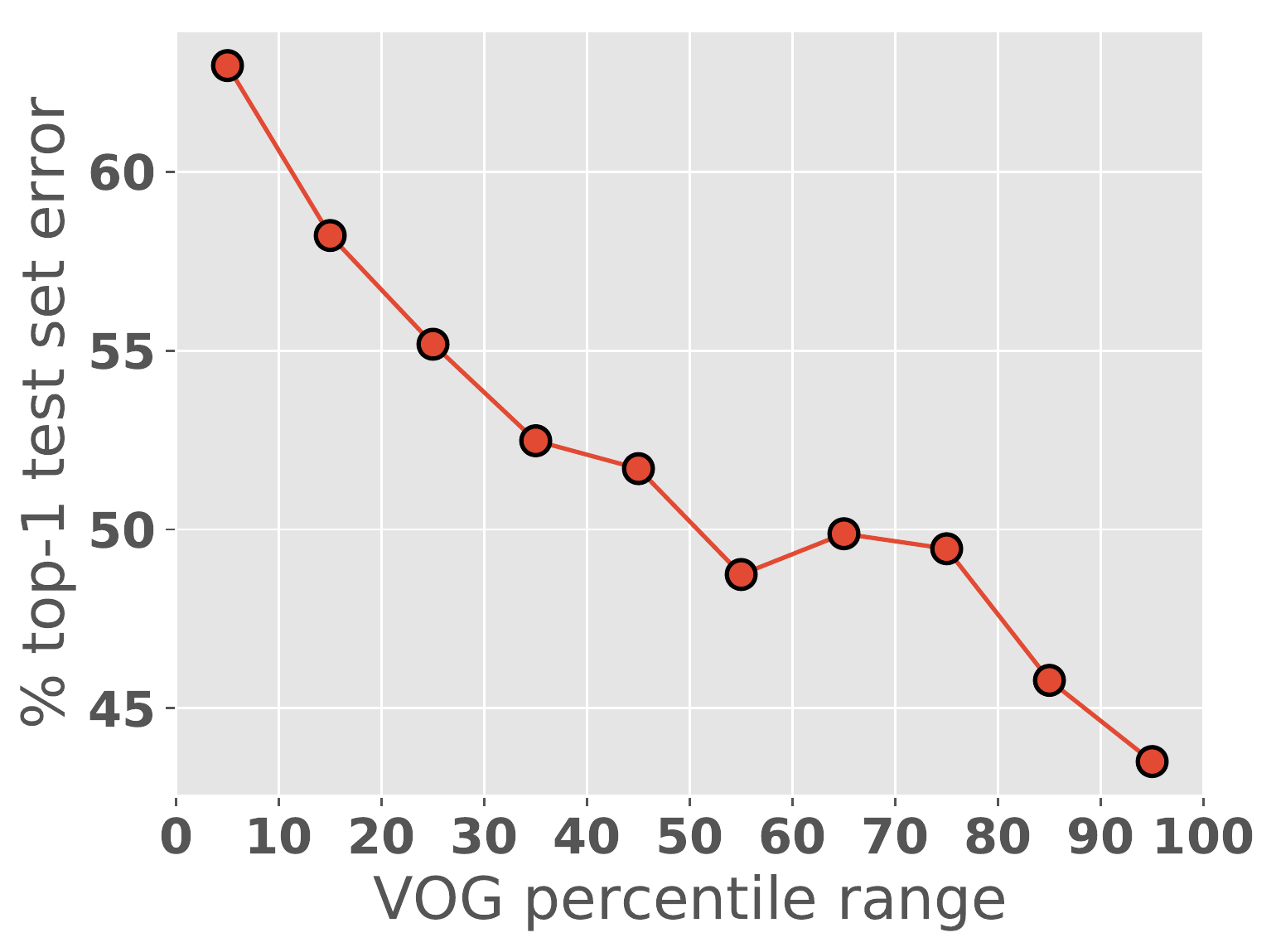}
        \caption{Early-stage training}
        \label{fig:imagenet_early}
	\end{subfigure}
	\begin{subfigure}{0.49\linewidth}
		\centering
    	\includegraphics[width=0.81\linewidth]{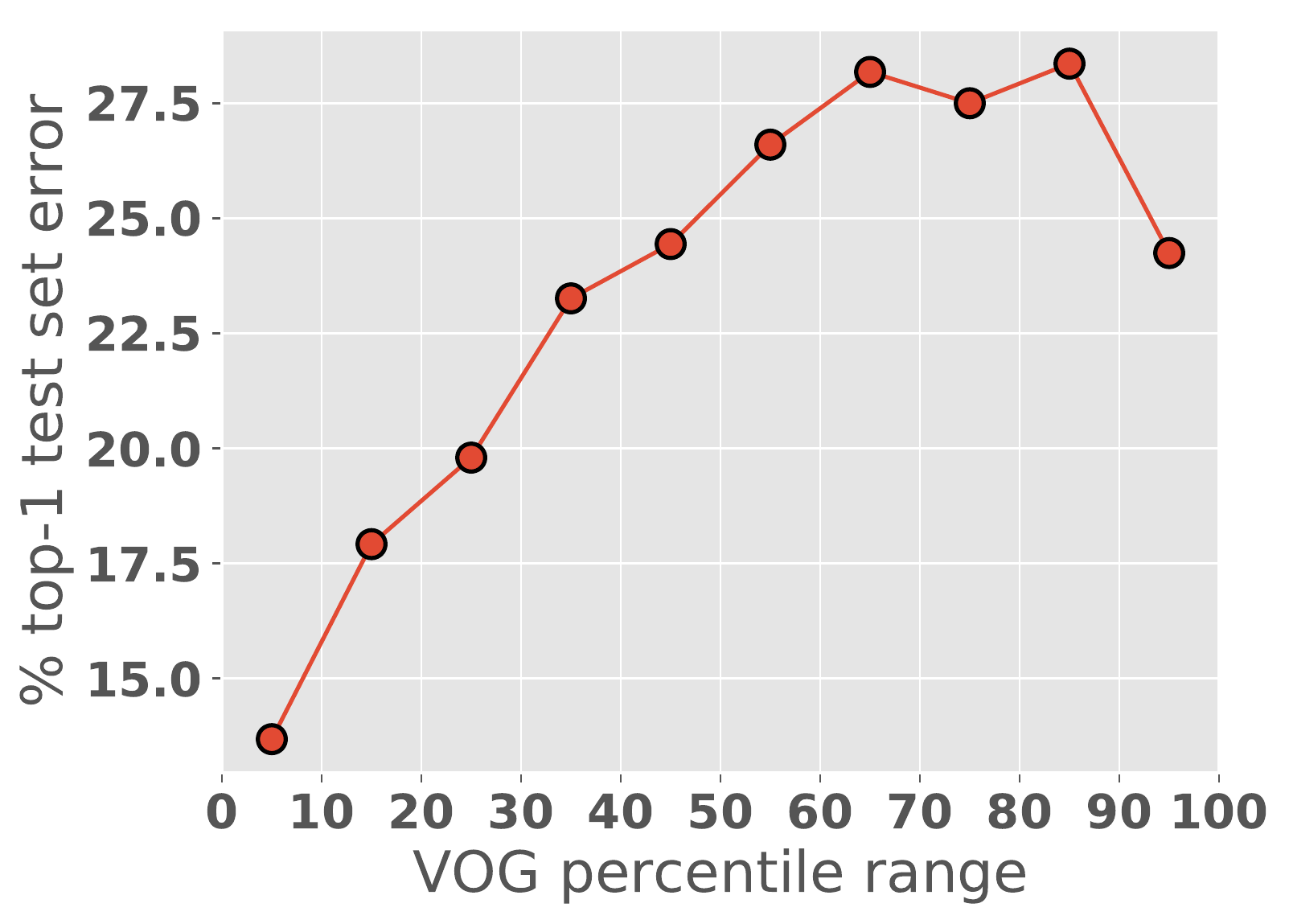}
		\caption{Late-stage training}
		\label{fig:imagenet_late}
	\end{subfigure}
    \vskip -0.1in
	\caption{
	    The mean top-1 test set error (y-axis) for the examples thresholded by VoG score percentile (x-axis) in ImageNet validation set.
	    The Early (a) and Late (b) stage VoG analysis shows inverse behavior where the role of VoG flips as the training progresses.
	}
	\label{fig:imagenet_error_plot}
	\end{sc}
\end{figure*}

\xhdr{4) VoG as an unsupervised auditing tool} Many auditing tools used to evaluate and understand possible model bias require the presence of labels for protected attributes and underlying variables. However, this is highly infeasible in real-world settings \cite{veale2017fairer}. For image and language datasets, the high dimensionality of the problem makes it hard to identify a priori what underlying variables one needs to be aware of. Even acquiring the labels for a limited number of attributes protected by law (gender, race) is expensive and/or may be perceived as intrusive, leading to noisy or incomplete labels \cite{HOOKER2021100241,mckane2020}. This means that ranking techniques which do not require labels at test time are very valuable. 

One key advantage of VoG is that we show it continues to produce a reliable ranking even when the gradients are computed \textit{w.r.t.} the predicted label. In Fig.~\ref{fig:vog_predicted}, we include the top and bottom 25 VoG ImageNet test images using predicted labels from the model. Finally, we also computed the mean test-error for the predicted VoG distribution, and find that it also effectively discriminates between top-10 and bottom-10 examples, respectively (Fig.~\ref{fig:error_rate_predicted}).

\begin{figure*}[ht!]
	\centering
\begin{sc}
	\begin{subfigure}{0.45\linewidth}
		\centering
    	\includegraphics[width=0.81\linewidth]{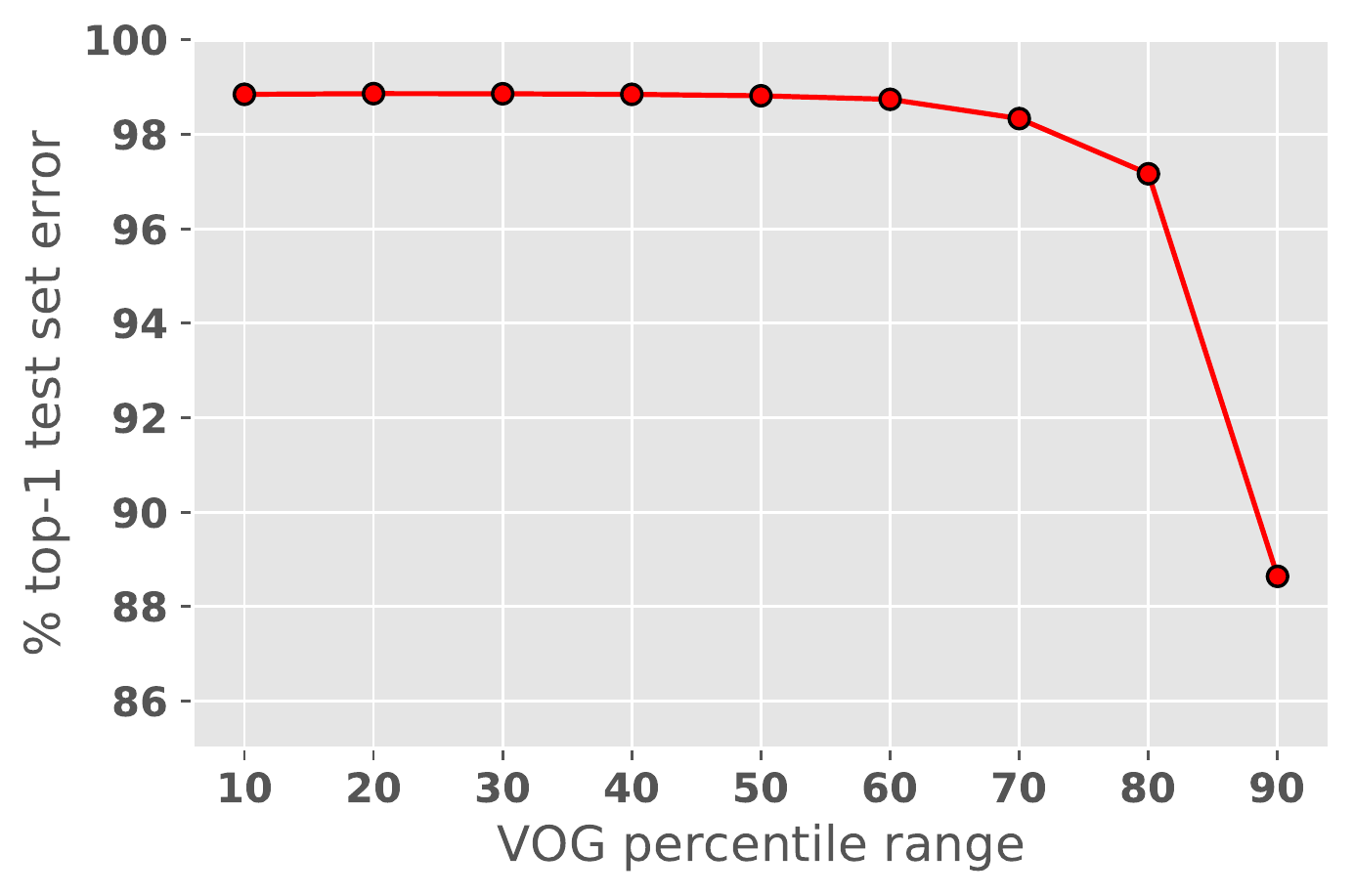}
    	\caption{Early-stage training}
        \label{fig:cifar100_tse_early}
	\end{subfigure}
	\begin{subfigure}{0.45\linewidth}
		\centering
    	\includegraphics[width=0.81\linewidth]{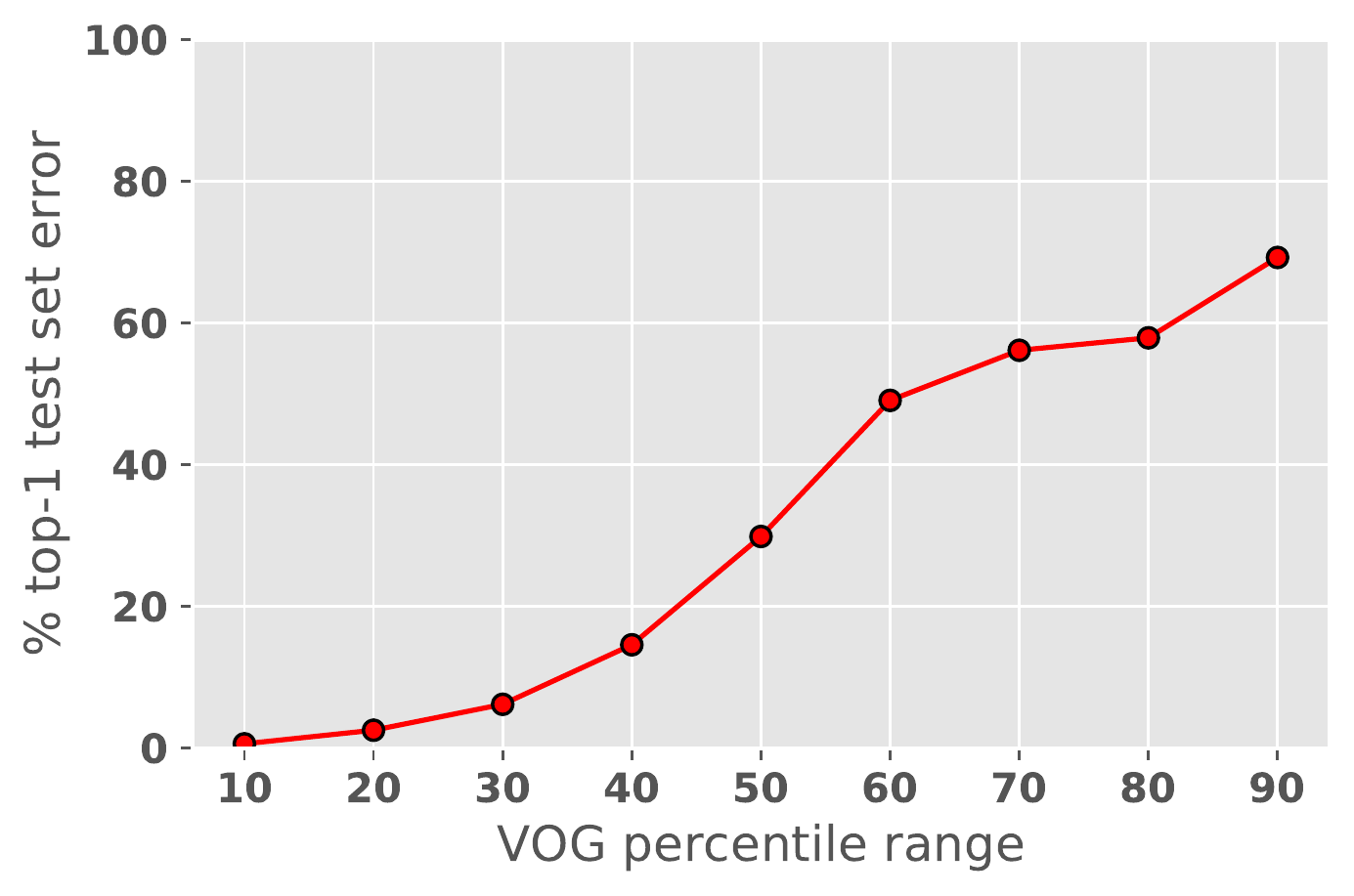}
    	\caption{Late-stage training}
        \label{fig:cifar100_tse_late}
	\end{subfigure}
	\vskip -0.1in
	\caption{
	    The mean top-1 test set error (y-axis) for the exemplars thresholded by VoG score percentile (x-axis) in Cifar-100 testing set.
	    The early (a) and late (b) stage VoG analysis shows inverse behavior where the role of VoG flips as the training progresses. Results for Cifar-10 are shown in Appendix Fig.~\ref{fig:cifar10_tse}.
	}
	\label{fig:cifar100_tse}
	\end{sc}
\end{figure*}

\xhdr{5) VoG understands early and late training dynamics}
\label{sec:vog_different_training_points}
Recent works have shown that there are distinct stages to training in deep neural networks \cite{achille2018critical, jiang2020characterizing, mangalam2019deep, faghri2020study}. To this end, we investigate whether VoG rankings are sensitive to the stage of the training process. We compute VoG separately for two different stages of the training process: (i) the \emph{Early}-stage (first three epochs) and (ii) the \emph{Late}-stage (last three epochs). We plot VoG scores against the test set error at each decile in early- and late-stage and find a flipping behavior across all datasets and networks (Fig.~\ref{fig:imagenet_error_plot} for ImageNet, Fig.~\ref{fig:cifar100_tse} for Cifar-100, and Fig.~\ref{fig:cifar10_tse} for Cifar-10). In the early training stage, samples having higher VoG scores have a lower average error rate as the gradient updates hinge on easy examples. This phenomenon reverses during the late-stage of the training, where, across all datasets, high VoG scores in the late-stage have the highest error rates as updates to the challenging examples dominate the computation of variance. Further, we note a noticeable visual difference between the image ranking computed for \textit{early-} and \textit{late}-stages of training. As seen in Fig.~\ref{fig:cifar100_vog_images}, for some classes such as \textit{apple}, it appears that VoG scores also capture the network's color bias during the \textit{early} training stage, where images with the lowest VoG scores over-index on red-colored apples.

\section{Relationship between VoG Scores and Memorized/OoD Examples}
\label{sec:vog_misclassify}
Recent works have highlighted that DNNs produce uncalibrated output probabilities that cannot be interpreted as a measure of certainty \cite{guo2017calibration,hendrycks2016baseline, kendall2017uncertainties, Lakshminarayanan2017}. To this end, we argue that if VoG is a reliable auditing tool, it should capture model uncertainty even when it's not reflected in the output probabilities. We consider VoG rankings on a task where the network produces highly confident predictions for incorrect/out-of-distribution inputs and evaluate VoG on two separate tasks: (1) identifying examples memorized by the model and (2) detecting out-of-distribution examples.

\subsection{Surfacing examples that require memorization}
Overparameterized networks have been shown to achieve zero training error by memorizing examples \cite{feldman2020does,hooker2020characterising,zhang2016understanding}. We explore whether VoG can distinguish between examples that require memorization and the rest of the dataset. To do this, we replicate the general experiment setup of Zhang et al. \cite{zhang2016understanding} and replace $20 \%$ of all labels in the training set with randomly shuffled labels. We re-train the model from random initialization and compute VoG scores \emph{across training} for all examples in the training set. Our network achieves $0\%$ training error which would only be possible given successful memorization of the noisy examples with shuffled labels. We now answer the question: \emph{Is VoG able to discriminate between these memorized examples and the rest of the dataset?}

We perform a two-sample $t$-test with unequal variances \cite{welch1947generalization} and show that this difference is statistically significant at a $p$-value of $0.001$, \ie shuffled labels have a different VoG distribution than the non-shuffled dataset. Intuitively, the two-sample $t$-test produces a $p$-value that can be used to decide whether there is evidence of a significant difference between the two distributions of VoG scores. The $p$-value represents the probability that the difference between the sample means is large, \ie the smaller the $p$-value, the stronger is the evidence that the two populations have different means. For both Cifar-10 and Cifar-100, we find a statistically significant difference in VoG scores for each population ($p$-value is $<0.001$), which shows that VoG is discriminative at distinguishing between memorized and non-memorized examples. We include more details about the statistical testing in Sec.~\ref{sec:app_stat_test}.

\subsection{Out-of-Distribution detection}
\label{sec:ood}
We have already established that VoG is very effective at distinguishing between easy and challenging examples (Fig.~\ref{fig:bar_error_rates}). Here, we ask whether this makes VoG an effective out of distribution (OoD) detection tool. It also gives us a setting in which to compare VoG as a ranking mechanism to other methods

Ruff et al. \cite{ruff2021unifying} benchmark a variety of OoD detection techniques on MNIST-C \cite{mu2019mnist}. For completeness, we replicate this precise setup by using a trained LeNet model and evaluate VoG on MNIST-C against 9 other methods\cite{rosenblatt1956remarks,rousseeuw2005robust,ruff2019deep,chalapathy2017robust,tax2004support,pearson1901liii,lee2018simple,kim2019rapp,schlegl2017unsupervised}.

\xhdr{Evaluation metrics} We evaluate OoD detection performance using the following metrics:

\xhdr{i) AUROC} The Area Under the Receiver Operator Characteristic (AUROC) curve can be interpreted as the probability that a positive example is assigned a higher detection score than a negative example \cite{fawcett2006introduction}.

\xhdr{ii) AUPR (In)} The Area Under the Precision Recall (AUPR) curve computes the precision-recall pairs for different probability thresholds by considering the in-distribution examples as the positive class.

\xhdr{iii) AUPR (Out)} AUPR (Out) is AUPR as described above, but calculated considering the OoD examples as the positive class. We treat this outlier class as positive by multiplying the VoG scores by  $-1$ and labelling them positive when calculating AUPR (Out).

\begin{table}[h]
  \centering\small
  \caption{Comparison of VoG to 9 existing OoD detection methods. Shown are average values of metrics and standard deviations across 15 corruptions in the MNIST-C datasets. Arrows ($\uparrow$) indicate the direction of better metric performance. VoG outperforms most baselines by a large margin.}
  \vskip -0.1in
  \begin{tabular}{lcc}
    \toprule
    OoD methods     & AUROC ($\uparrow$) & AUPR OUT ($\uparrow$)\\
    \midrule
    KDE\cite{rosenblatt1956remarks} & 57.46\std{32.09}    & 62.56\std{24.16}   \\
    MVE\cite{rousseeuw2005robust} & 62.84\std{21.92}    & 61.42\std{19.1}   \\
    DOCC\cite{ruff2019deep} & 69.16\std{28.35}    & 70.37\std{23.25}   \\
    kPCA\cite{chalapathy2017robust} & 72.12\std{31.00}.   & 75.39\std{26.37}   \\
    SVDD\cite{tax2004support} & 74.01\std{21.39}    & 73.33\std{21.98}   \\
    PCA\cite{pearson1901liii} & 77.71\std{30.90}    & 80.86\std{25.2}   \\
    Gaussian\cite{lee2018simple}  & 80.57\std{29.71}    & 84.51\std{22.62}   \\
    \textbf{VoG} & 85.42\std{10.28}.   & 84.96\std{9.61}   \\
    AE\cite{kim2019rapp} & 89.89\std{18.52}    & 89.99\std{18.19}   \\
    AEGAN\cite{schlegl2017unsupervised} & 95.93\std{7.90}.   & 95.40\std{9.46}   \\
    \bottomrule
  \end{tabular}
  \label{tab:mnistc}
\end{table}

\xhdr{Findings} In Table~\ref{tab:mnistc}, we observe that VoG outperforms all methods except AutoEncoders (AE) and AutoEncoder GAN (AEGAN). 
In stark contrast to VoG, AE and AEGAN require complex training of auxiliary models and do not feasibly scale beyond small-scale datasets like MNIST. Given these limitations, VoG remains a valuable and scalable OoD detection method as it can be used for large-scale datasets (\eg ImageNet) and networks (\eg ResNet-50). Unlike generative models, VoG does not require an uncorrupted training dataset for learning image distributions.
Further, VoG only leverages data from training itself, is computed from checkpoints already stored over the course of training, and does not require the true label to rank.

\section{Related Work}\label{sec:related_work} 
Our work proposes a method to rank training and testing data by estimating example difficulty. Given the size of current datasets, this can be a powerful interpretability tool to isolate a tractable subset of examples for human-in-the-loop auditing and aid in curriculum learning~\cite{bengio2009curriculum} or distinguishing between sources of uncertainty ~\cite{hu2021does,dsouza2021tale}. While prior works have proposed different notions of what subset merits surfacing, introduced the concept of prototypes and quintessential examples in the dataset, but did not focus on large-scale deep neural networks models~\cite{zhang1992selecting,bien2011prototype,kim2014bayesian,kim2016examples,chang2017active}.

Unlike previous works, we propose a measure that can be extended to rank the entire dataset by estimating example difficulty (rather than surfacing a prototypical subset). In addition, VoG is far more efficient than other global rankings like \cite{koh2017understanding} and \cite{harutyunyan2021estimating}. 

VoG also does not require modifying the architecture or making any assumptions about the statistics of the input distribution. In particular, works such as \cite{kim2016examples} require assumptions about the statistics of the input distribution and \cite{li2018deep} requires modifying the architecture to prefix an autoencoder to surface a set of prototypes, \cite{paul2021deep} leverages pruning of the model to identify difficult examples and \cite{baldock2021} requires the addition of an auxiliary k-nn model after each layer. 

\looseness=-1
Our work is complementary to recent works by \cite{jiang2020characterizing} that proposes a c-score to rank examples by aligning them with training instances, \cite{hooker2019compressed} that classifies examples as outliers according to sensitivity to varying model capacity, and \cite{carlini2019distribution} that considers different measures to isolate prototypes for ranking the entire dataset. We note that the c-score method proposed by \cite{jiang2020characterizing} is considerably more computationally intensive to compute than VoG as it requires training up to 20,000 network replications per dataset. Several of the prototype methods considered by \cite{carlini2019distribution} require training ensembles of models, as does the compression sensitivity measure proposed by \cite{hooker2019compressed}. Finally, our proposed VoG is both different in the formulation and can be computed using a small number of existing checkpoints saved over the course of training.

\section{Conclusion and Future Work}
\label{sec:conclusions}
In this work, we proposed VoG as a valuable and efficient way to rank data by difficulty and surface a tractable subset of the most challenging examples for human-in-the-loop auditing. High VoG samples are challenging to classify for  algorithm and surfaces clusters of images with distinct visual properties. Moreover, VoG is domain agnostic as it uses only the vanilla gradient explanation from the model, and can be used to rank both training and test examples. We show that it is also a useful unsupervised protocol, as it can effectively rank examples using the predicted label.


\newpage
\bibliography{main}
\bibliographystyle{ieee_fullname}

\clearpage
\appendix
\section{Appendix}
\subsection{Toy Experiment}
\looseness=-1
We generate the clusters for classification using \href{https://scikit-learn.org/stable/modules/generated/sklearn.datasets.make_blobs.html}{scikit-learn} and use a 90-10\% split for dividing the dataset into train and test set~\footnote{Code and datasets available at \url{https://github.com/chirag126/VOG.git}}. We train a linear Multiple Layer Perceptron network with a hidden layer of 10 neurons using Stochastic Gradient Descent optimizer for 15 epochs. We divided the training process into three epoch stages: (1) $Early$ [0, 5), (2) $Middle$ [5, 10), and (3) $Late$ stage [10, 15). The trained model achieves a 0\% test set error using a linear boundary (Fig.~\ref{fig:toy_decision_boundary}).

\subsection{Class Level Error Metrics and VoG} 
\begin{figure*}[h]
	\centering
	\begin{flushleft}
        \hspace{3.3cm}$r$=0.65~$p$-val=0.04\hspace{5.7cm}$r$=0.59~$p$-val=8.39e-11
    \end{flushleft}
    \vspace{-4mm}
	\begin{subfigure}{0.48\linewidth}
		\centering
    	\includegraphics[width=0.72\linewidth]{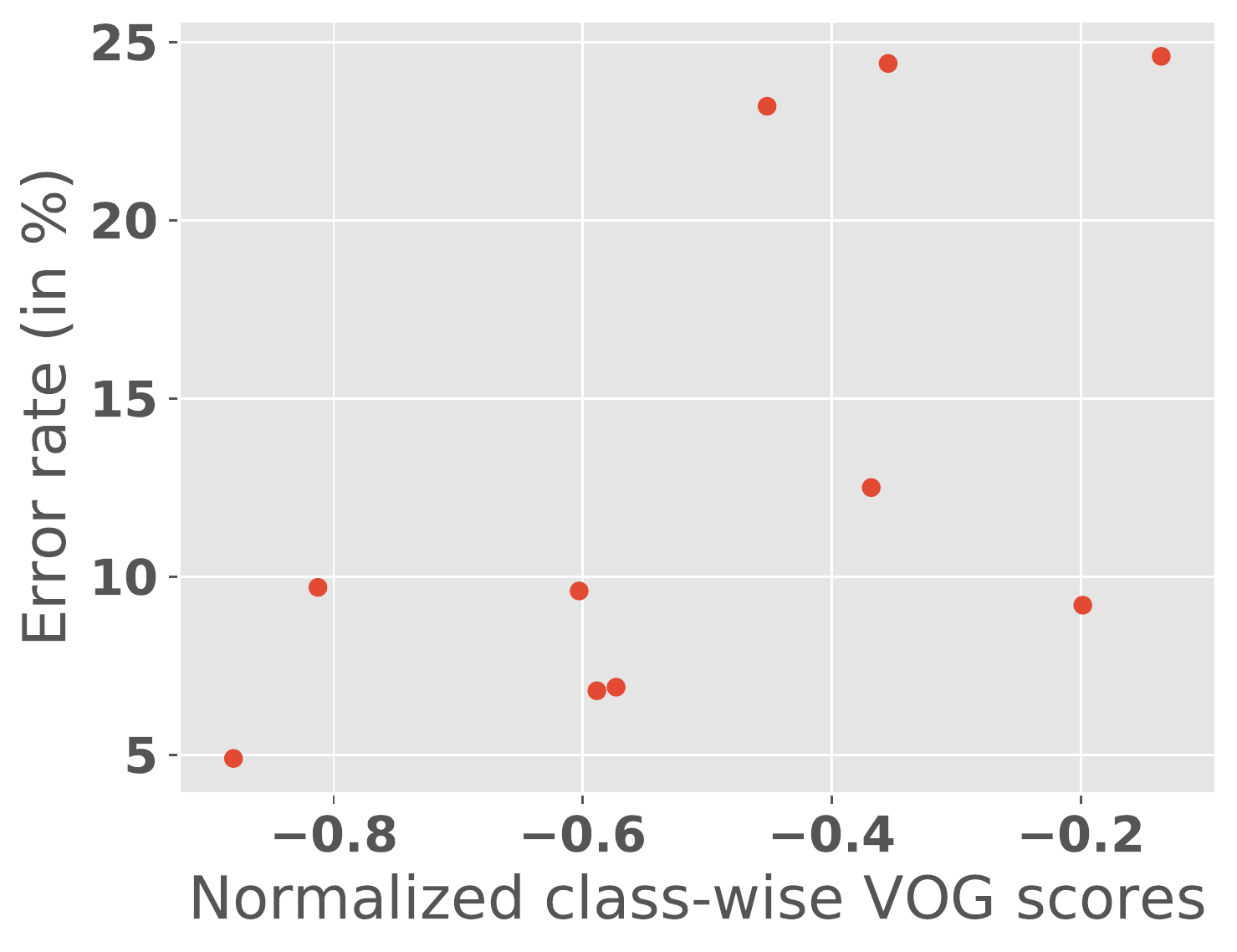}
    	\caption{Cifar-10}
        \label{fig:acc_vog_cifar10}
	\end{subfigure}
	\begin{subfigure}{0.49\linewidth}
		\centering
    	\includegraphics[width=0.72\linewidth]{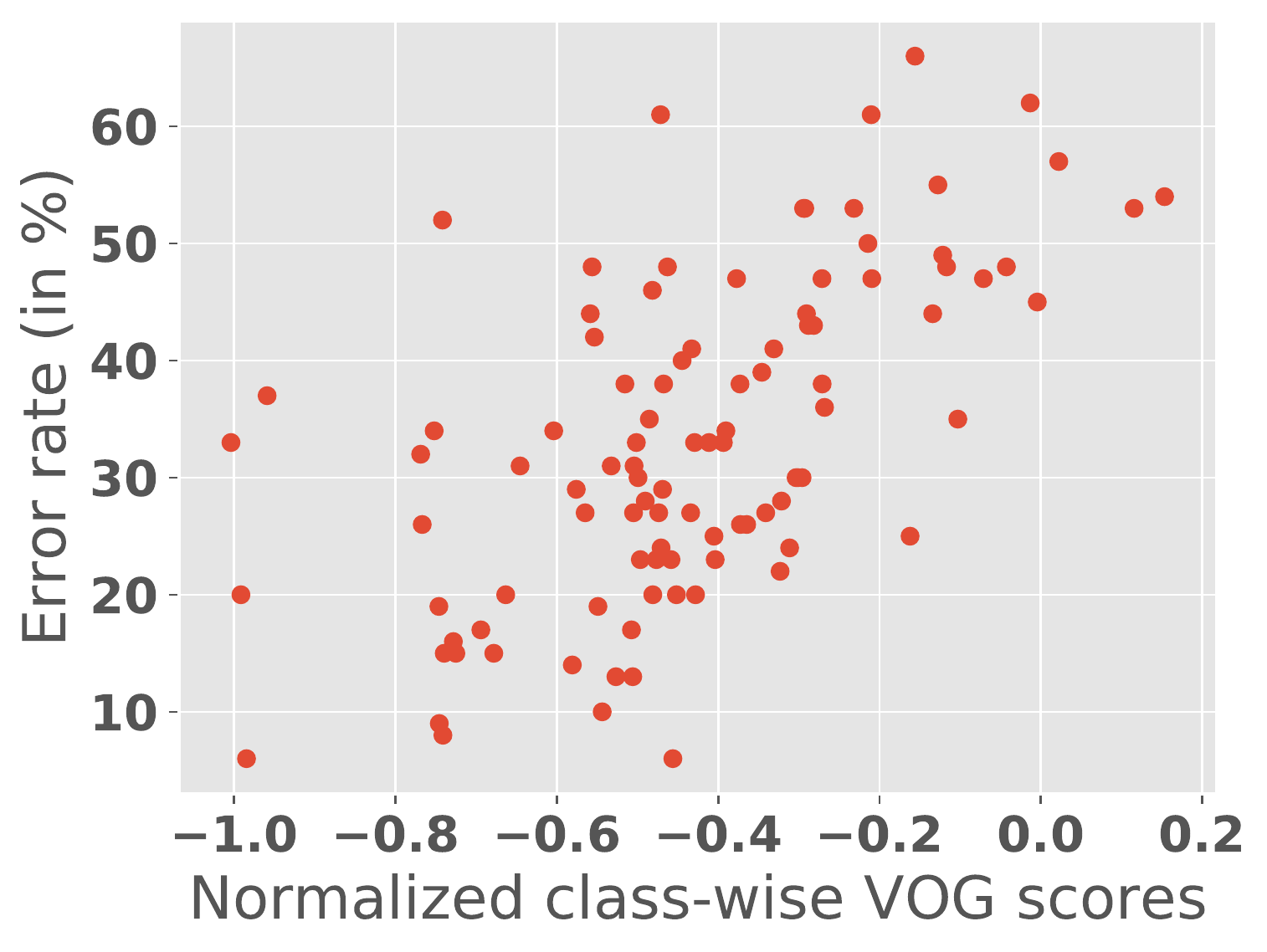}
    	\caption{Cifar-100}
        \label{fig:acc_vog_cifar100}
	\end{subfigure}
	\caption{
	    Plot of error rate (y-axis) against normalized class VoG scores for all classes (x-axis). There is a statistically significant positive correlation between class level error metrics and average  VoG score (alpha set at 0.05).
	}
	\label{fig:acc_vog}
\end{figure*}

\looseness=-1
Here, we explore whether VoG is able to capture class level differences in difficulty. We compute VoG scores for each image in the test set of Cifar-10 and Cifar-100 (both test sets have $10,000$ images). In Fig.~\ref{fig:acc_vog}, we plot the average absolute VoG score for each class against the false negative rate for each class. We find that there is a positive, albeit weak, correlation between the two, classes with higher VoG scores have higher mis-classification error rate. The correlation between these metrics is $0.65$ and $0.59$ for Cifar-10 and Cifar-100 respectively. Given that VoG is computed on a per-example level, we find it interesting that the aggregate average of VoG is able to capture class level differences in difficulty.

\begin{figure*}[h]
	\centering
	\begin{subfigure}{0.33\linewidth}
		\centering
    	\includegraphics[width=0.81\linewidth]{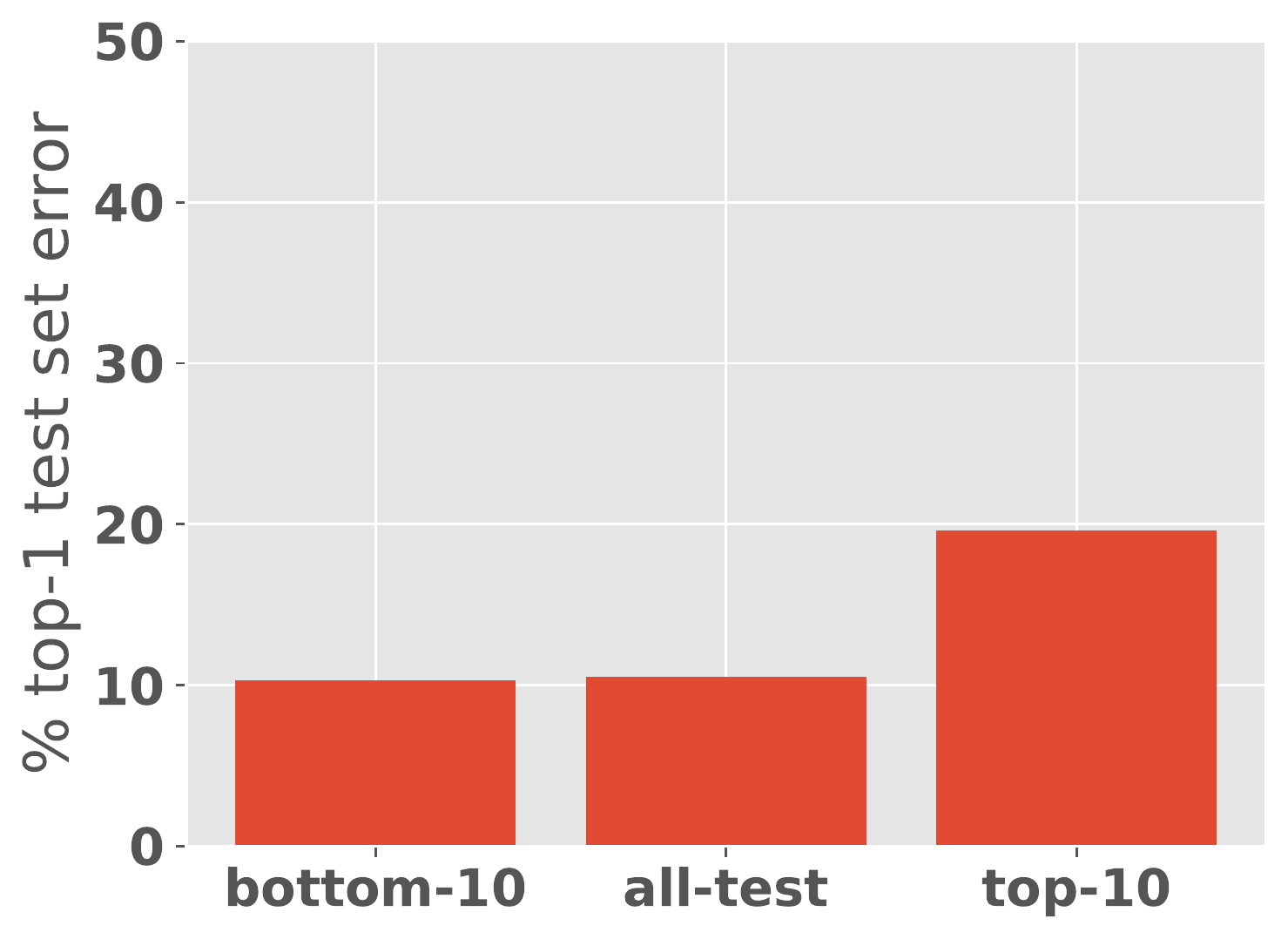}
    	\caption{Cifar-10}
        \label{fig:bar_cifar10}
	\end{subfigure}
	\begin{subfigure}{0.33\linewidth}
		\centering
    	\includegraphics[width=0.81\linewidth]{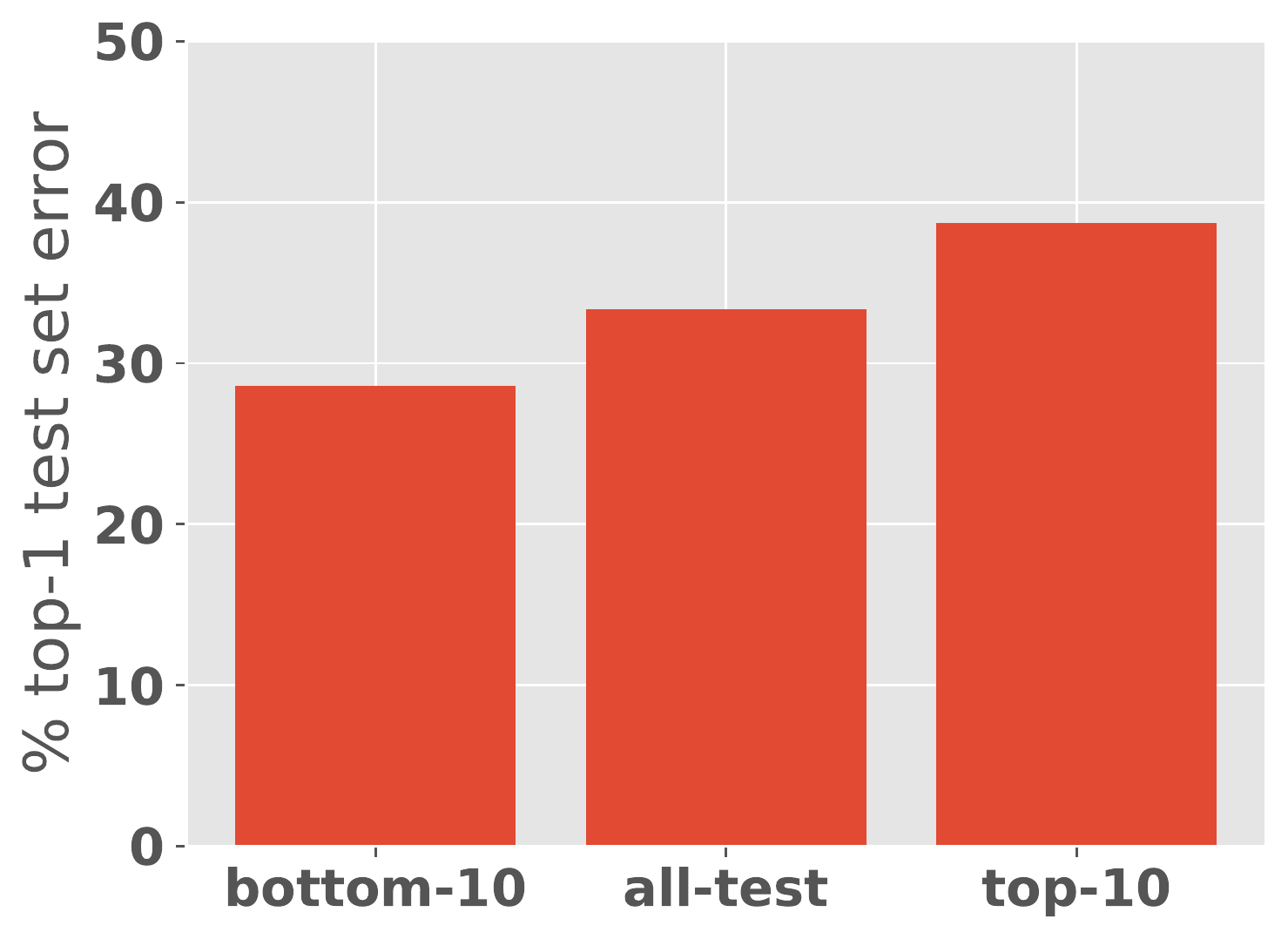}
    	\caption{Cifar-100}
        \label{fig:bar_cifar100}
	\end{subfigure}
	\begin{subfigure}{0.33\linewidth}
		\centering
    	\includegraphics[width=0.81\linewidth]{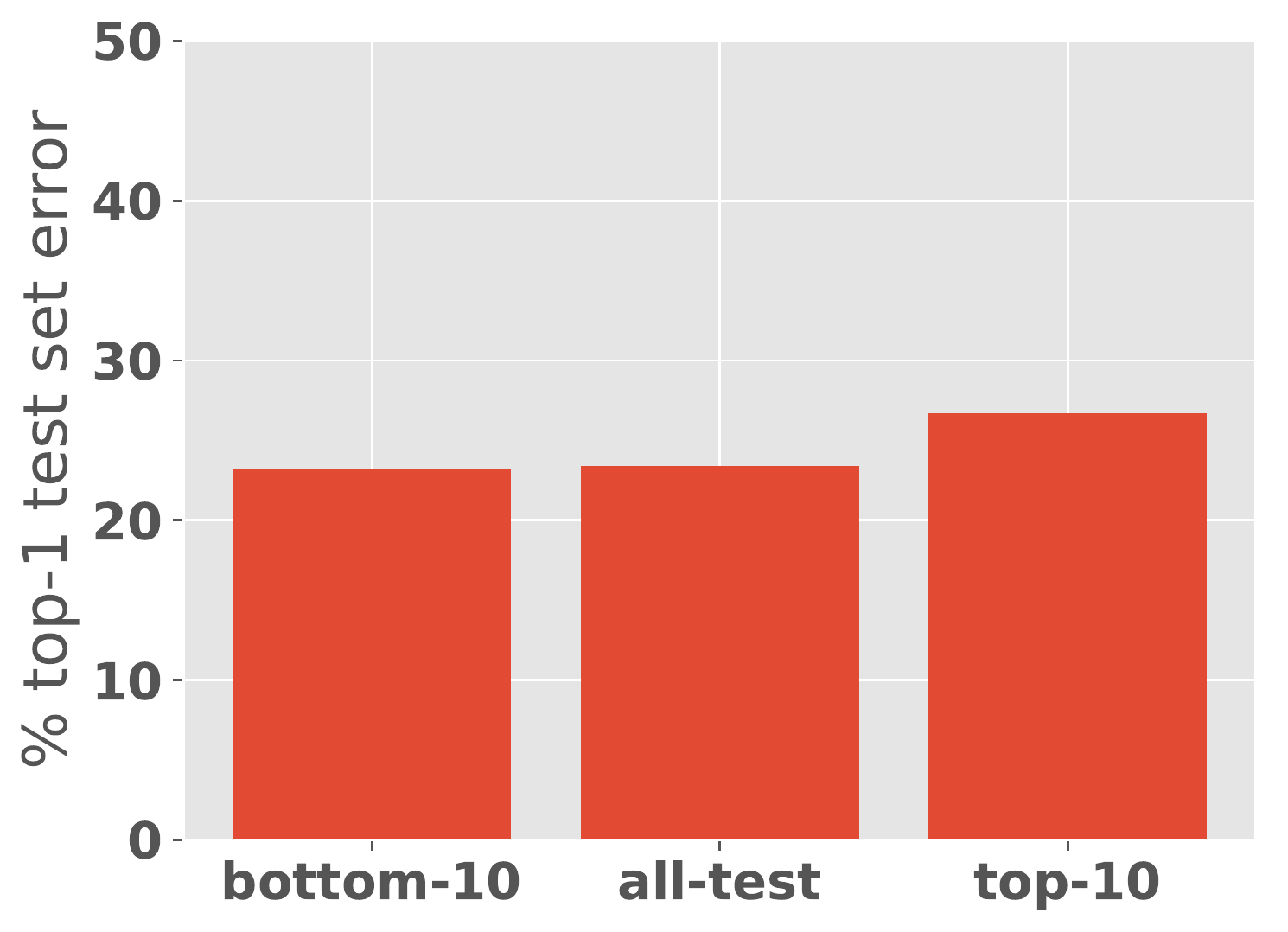}
    	\caption{ImageNet}
        \label{fig:bar_imagenet}
	\end{subfigure}
	\caption{
	    Bar plots showing the mean top-1 error rate (in $\%$) for three group of samples from (1) the subset of the test set with the bottom 10th percentile of VoG scores, (2) the complete testing dataset, and (3) the subset of the test set with the top 10th percentile of VoG scores. 
	}
	\label{fig:bar_error_rates}
\end{figure*}
\subsection{Statistical Significance of Memorization Experiments}
\label{sec:app_stat_test}
\looseness=-1
The two-sample $t$-test produces a $p$-value that can be used to decide whether there is evidence of a significant difference between the two distributions of VoG scores. The $p$-value represents the probability that the difference between the sample means is large, \ie smaller the $p$-value, stronger is the evidence that the two populations have different means. 

\textbf{Null Hypothesis}:~ $\mu_{1}=\mu_{2}$~~~~~\textbf{Alternative Hypothesis}:~ $\mu_{1} \neq \mu_{2}$

If the $p$-value is less than your significance level ($\alpha = 0.05$ in this experiment), you can reject the null hypothesis, \ie the difference between the two means is statistically significant. The details for the individual $t$-tests for Cifar-10 and Cifar-100 are given below:

\textbf{Cifar-10:}~The statistics for the samples in the correct and shuffled labels are:\\
Corrected labels:~ $\mu_{1}=0.62$; $\sigma_{1}=0.54$; $N_{1}=40000$ \\
Shuffled labels:~ $\mu_{2}=0.85$; $\sigma_{2}=0.75$; $N_{2}=10000$ \\ 
\underline{Result:}~$p$-value is $<0.001$~|~Reject Null Hypothesis (the two populations have different VoG means)

\textbf{Cifar-100:}~The statistics for the samples in the correct and shuffled labels are:\\
Corrected labels:~ $\mu_{1}=0.54$; $\sigma_{1}=0.46$; $N_{1}=40000$ \\
Shuffled labels:~ $\mu_{2}=0.82$; $\sigma_{2}=0.71$; $N_{2}=10000$ \\
\underline{Result:}~$p$-value is $<0.001$ | Reject Null Hypothesis (the two populations have different VoG means)
\subsection{Early training dynamics of Deep Neural Networks}
\begin{figure*}
	\centering
	\begin{subfigure}{0.49\linewidth}
		\centering
    	\includegraphics[width=0.81\linewidth]{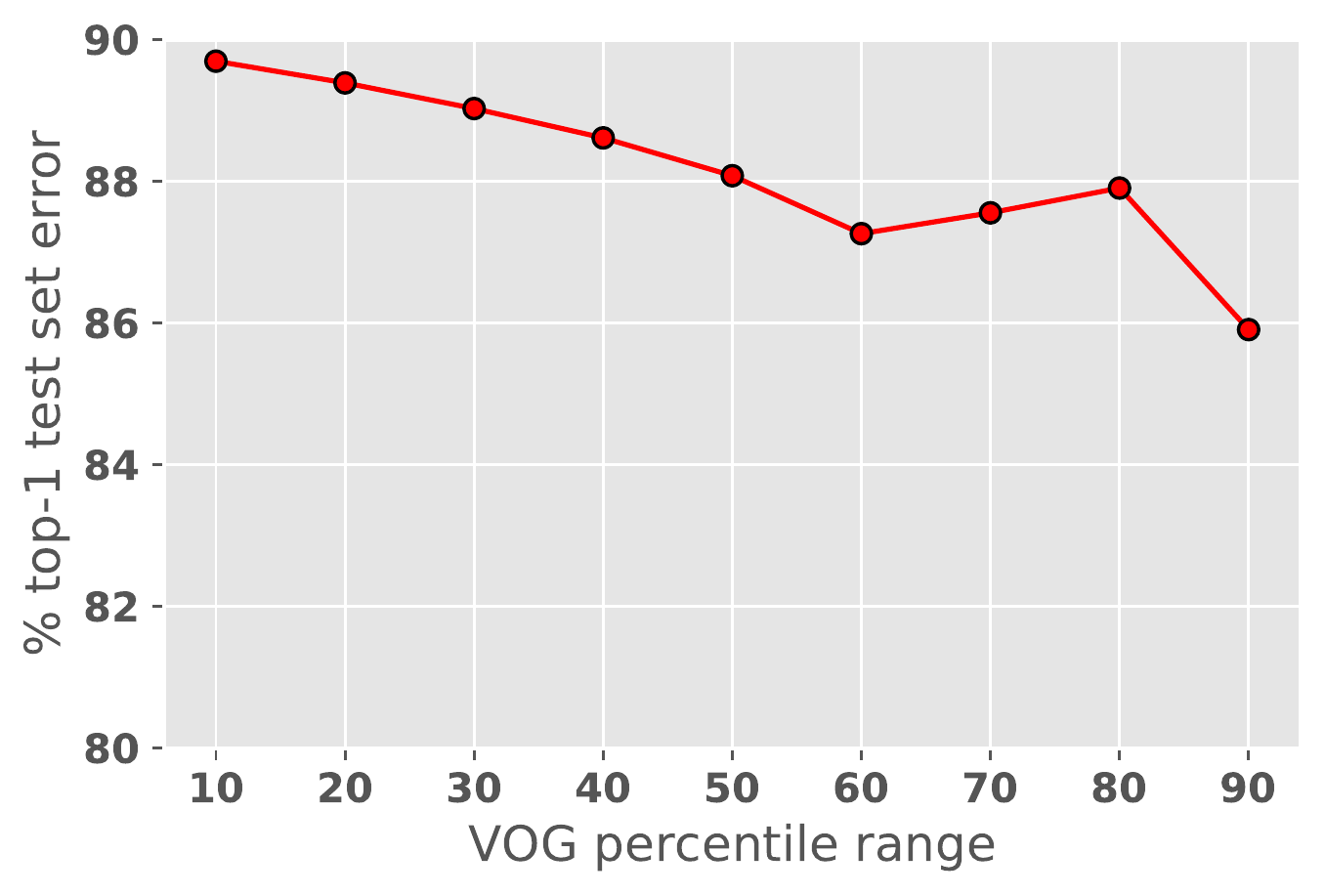}
    	\caption{Early-stage training}
        \label{fig:cifar10_tse_early}
	\end{subfigure}
	\begin{subfigure}{0.49\linewidth}
		\centering
    	\includegraphics[width=0.81\linewidth]{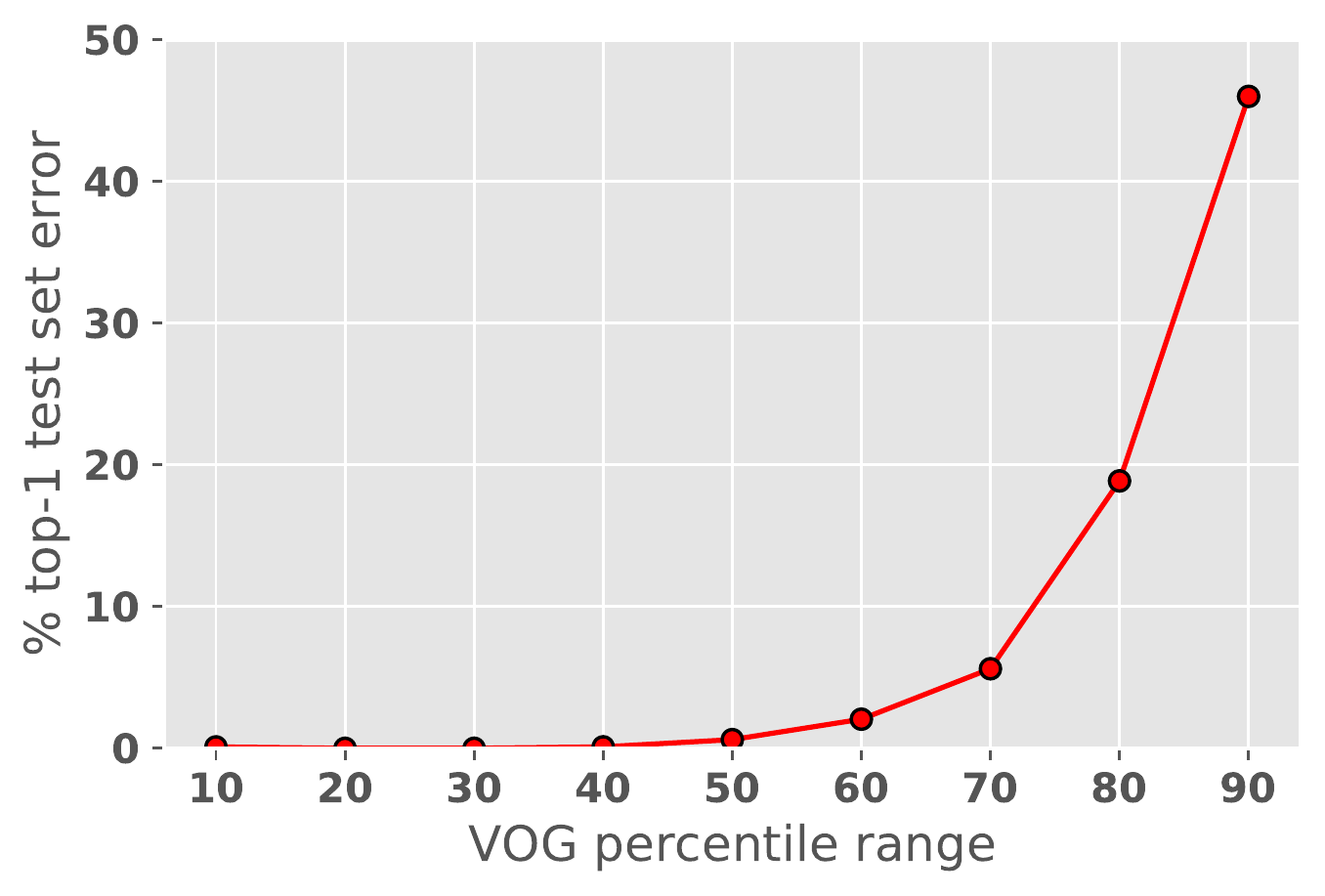}
    	\caption{Late-stage training}
        \label{fig:cifar10_tse_late}
	\end{subfigure}
	\caption{
	    The mean top-1 test set error (y-axis) for the exemplars thresholded by VoG score percentile (x-axis) in Cifar-10 testing set.
	    The Early (a) and Late (b) stage VoG analysis shows inverse behavior where the role of VoG flips as the training progresses.
	}
	\label{fig:cifar10_tse}
\end{figure*}

Following Sec.~\ref{sec:vog_different_training_points}, we plot the relationship between VoG and error rate of the testing dataset for Cifar-10 and Cifar-100.
As in ImageNet, we observe a \textit{flipping} trend between the early and late stages for both datasets (Figs.~\ref{fig:cifar100_tse},\ref{fig:cifar10_tse}).
We find that for easier datasets like Cifar, this point is only seen on using a lower learning rate (1e-3 in our experiments) for the early training stages.
\subsection{Detection of Distribution Shifts}
\begin{figure*}
    \begin{sc}
        \begin{subfigure}{0.49\linewidth}
            \centering
    	    \includegraphics[width=0.81\linewidth]{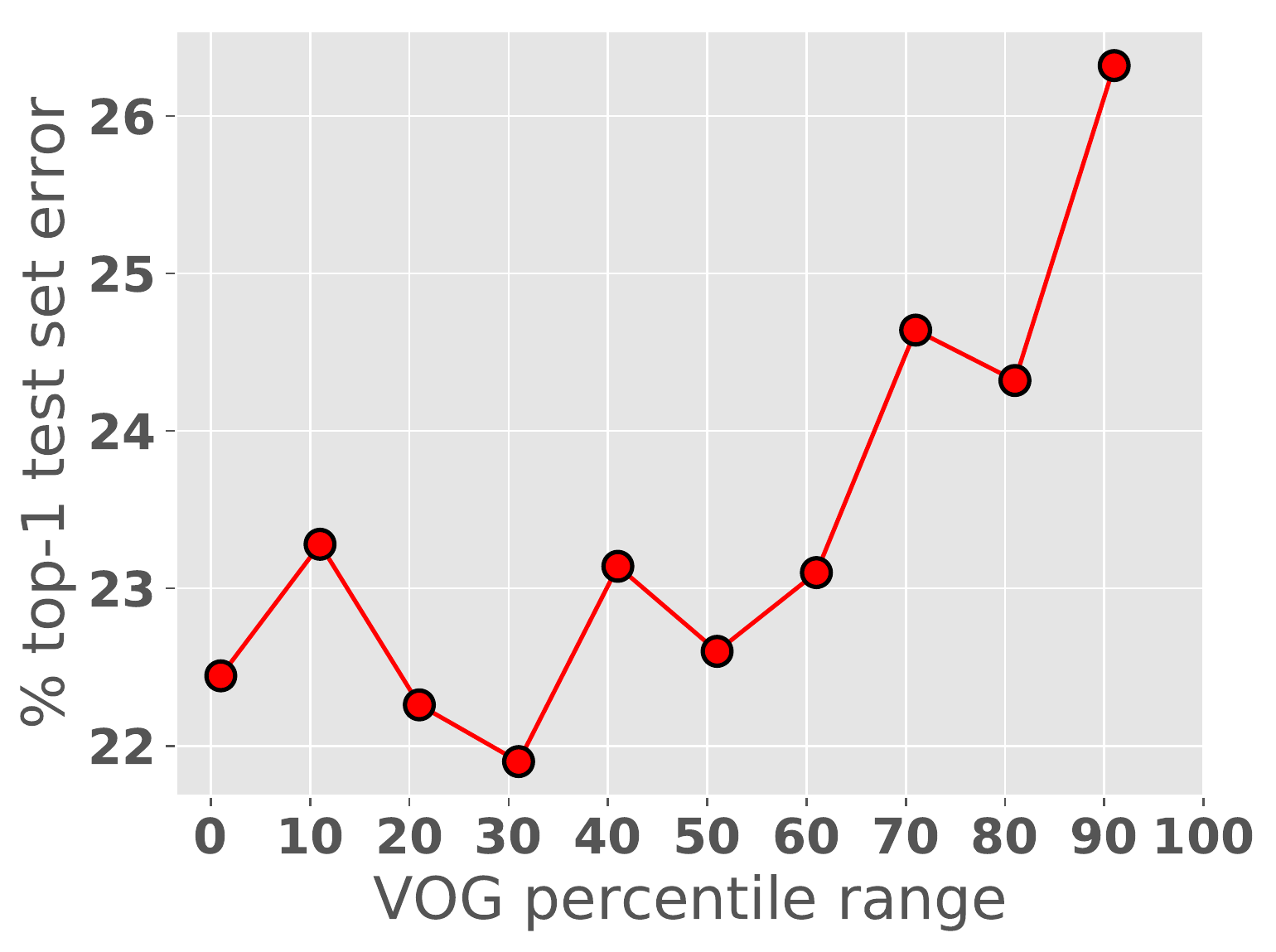}
    	    \caption{}
    	    \label{fig:error_rate_predicted}
        \end{subfigure}
        \begin{subfigure}{0.49\linewidth}
        	\centering
            \includegraphics[width=0.81\linewidth]{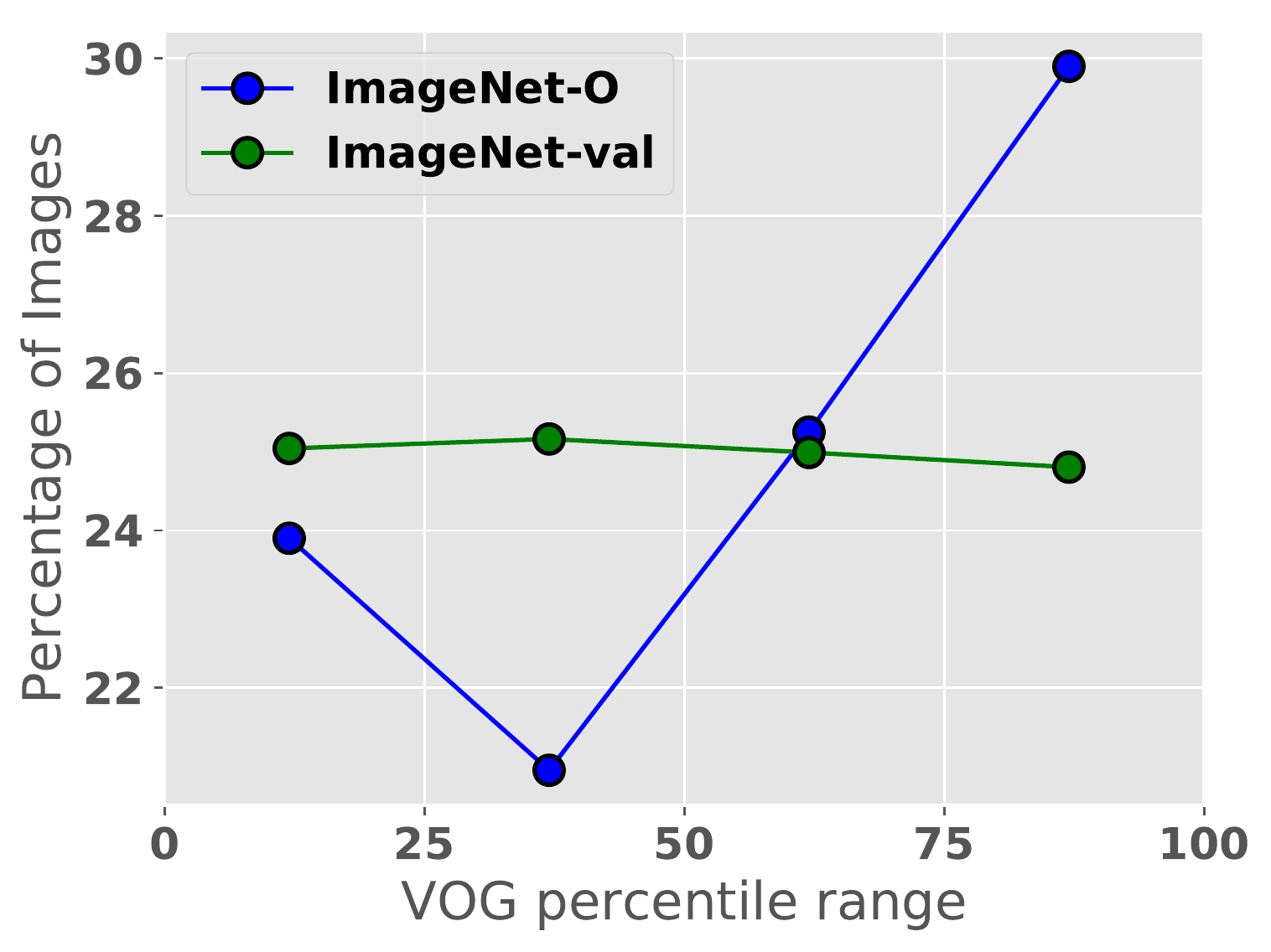}
            \caption{}
        	\label{fig:imagenet_o}
        \end{subfigure}
	\end{sc}
	\caption{\textbf{Left:~} VoG is a valuable unsupervised tool as it can be computed using either the predicted/true label. We observe that misclassification increases with an increase in VoG scores. Across ImageNet, we observe that VoG calculated for the predicted labels follows the same trend as Fig.~\ref{fig:vog_predicted}, where the top-10 percentile VoG scores have the highest error rate. \textbf{Right:~} Number of ImageNet-O images across different VoG percentiles. We find that higher percentiles of VoG are significantly more likely to over-index on these OoD images.
	} 
\end{figure*}

We consider ImageNet-O \cite{hendrycks2021natural}, an open source curated out-of-distribution (OoD) dataset designed to fool classifiers. ImageNet-O consists of images that are not included in the original $1000$ ImageNet classes. These images were selected with the goal of producing high confidence incorrect ImageNet-1K predictions of labels from within the training distribution. We are interested in understanding if VoG can correctly rank ImageNet-O examples as being atypical or OoD and expect to observe that ImageNet-O examples would be over-represented in top percentiles of VoG scores.
In Fig.~\ref{fig:imagenet_o}, we observe that the 
percentage of ImageNet-O images are relatively over-represented at high levels of VoG, with 30\% of all images in the top-25th percentile vs 24\% in the bottom 25th percentile.

\subsection{Out-of-Distribution Detection (OoD) Datasets and Model Architectures}
Here, we carry out additional experiments to measure the effectiveness of VoG to detect OoD data. We run experiments using three DNN architectures: ResNet-18~\cite{he2016deep}, DenseNet~\cite{huang2017densely} and WideResNet \cite{zagoruyko2016wide}, and benchmark against Maximum Softmax Probability (MSP) \cite{hendrycks2016baseline}, which is widely considered a strong baseline in OoD detection \cite{hendrycks2016baseline,2020arXiv200406100H}
We follow the setup in \cite{hendrycks2016baseline} by setting all test set examples in CIFAR-10 as in-distribution (positive).
For OoD examples (negative), we benchmark across four datasets: CIFAR-100, iSUN \cite{liang2018enhancing}, TinyImageNet (Resize) \cite{liang2018enhancing}, LSUN (Resize) \cite{liang2018enhancing}, and Gaussian Noise. The Gaussian dataset was generated as described in \cite{liang2018enhancing}, with $\mathcal{N}(0.5, 1)$. For the various ablations, the size of the OoD dataset can be seen in Table~\ref{tab:dataset_size}.

\begin{table*}
\centering
    \begin{sc}
    \caption{Number of images for each of the OoD dataset used in our OoD detection experiments.}
    \label{tab:dataset_size}
    \begin{tabular}{lc}
        \hline
        Dataset & Dataset size \\\hline
        Cifar-100 & 10000 \\
        Gaussian & 10000 \\
        iSUN & 8920 \\
        Tiny-ImageNet-Resize & 9810 \\
        LSUN-Resize & 10000\\\hline
    \end{tabular}
    \end{sc}
\end{table*}

\xhdr{Findings} From Table~\ref{tab:baseline}, we observe that VoG is a valuable ranking for OoD detection and improves upon state-of-the-art uncertainty measures for many different tasks. On average, VoG outperforms MSP by large margins  with a mean gain of 2.62\% in AUROC, 2.33\% in AUPR/In, and 2.47\% in AUPR/Out across all three architectures and five datasets.

\begin{table*}
\centering
\begin{sc}
\setlength{\tabcolsep}{1.8pt}
\caption{Baseline comparison between VoG and Max Softmax Probability (MSP) for different models trained on Cifar-10.
VoG is able to detect, both, In- and Out-Of-Distribution (OoD) samples with higher precision across different real-world datasets.
For each row, values in \textbf{bold} represents superior performance.}
\begin{tabular}{llcccc}
\hline
Model & In- / Out-of-Distribution & Metrics & AUROC /Base & \makecell{AUPR \\ In /Base} & \makecell{AUPR \\ Out/Base} \\\hline
 & C-10/C-100 & \begin{tabular}[c]{@{}c@{}}MSP\\ VoG\end{tabular} & \multicolumn{1}{c}{\begin{tabular}[c]{@{}c@{}}80.9/50\\ \textbf{89}/50\end{tabular}} & \multicolumn{1}{c}{\begin{tabular}[c]{@{}c@{}}83.4/50\\ \textbf{90.5}/50\end{tabular}} & \multicolumn{1}{c}{\begin{tabular}[c]{@{}c@{}}75.4/50\\ \textbf{87.3}/50\end{tabular}} \\
 \multirow{5}{*}{W-RN-28-10} & C-10/Gaussian & \begin{tabular}[c]{@{}c@{}}MSP\\ VoG\end{tabular} & \multicolumn{1}{c}{\begin{tabular}[c]{@{}c@{}}78.1/50\\ \textbf{88.2}/50\end{tabular}} & \multicolumn{1}{c}{\begin{tabular}[c]{@{}c@{}}84.6/50\\ \textbf{91.6}/50\end{tabular}} & \multicolumn{1}{c}{\begin{tabular}[c]{@{}c@{}}66.4/50\\ \textbf{80.6}/50\end{tabular}} \\
 & C-10/iSUN & \begin{tabular}[c]{@{}c@{}}MSP\\ VoG\end{tabular} & \multicolumn{1}{c}{\begin{tabular}[c]{@{}c@{}}87.8/50\\ \textbf{93.3}/50\end{tabular}} & \multicolumn{1}{c}{\begin{tabular}[c]{@{}c@{}}90.7/52.8\\ \textbf{95.3}/52.8\end{tabular}} & \multicolumn{1}{c}{\begin{tabular}[c]{@{}c@{}}82.9/47.2\\ \textbf{89.4}/47.2\end{tabular}} \\
 & C-10/Tiny-ImageNet-Resize & \begin{tabular}[c]{@{}c@{}}MSP\\ VoG\end{tabular} & \multicolumn{1}{c}{\begin{tabular}[c]{@{}c@{}}88.4/50\\ \textbf{92.8}/50\end{tabular}} & \multicolumn{1}{c}{\begin{tabular}[c]{@{}c@{}}91/50.5\\ \textbf{94.3}/50.5\end{tabular}} & \multicolumn{1}{c}{\begin{tabular}[c]{@{}c@{}}83.4/49.5\\ \textbf{89.9}/49.5\end{tabular}} \\
 & C-10/LSUN-Resize & \begin{tabular}[c]{@{}c@{}}MSP\\ VoG\end{tabular} & \multicolumn{1}{c}{\begin{tabular}[c]{@{}c@{}}90.4/50\\ \textbf{93.5}/50\end{tabular}} & \multicolumn{1}{c}{\begin{tabular}[c]{@{}c@{}}92.7/50\\ \textbf{94.9}/50\end{tabular}} & \multicolumn{1}{c}{\begin{tabular}[c]{@{}c@{}}86.6/50\\ \textbf{90.8}/50\end{tabular}} \\\hline
 & C-10/C-100 & \begin{tabular}[c]{@{}c@{}}MSP\\ VoG\end{tabular} & \multicolumn{1}{c}{\begin{tabular}[c]{@{}c@{}}86.8/50\\ \textbf{87.6}/50\end{tabular}} & \multicolumn{1}{c}{\begin{tabular}[c]{@{}c@{}}89.7/50\\ \textbf{90}/50\end{tabular}} & \multicolumn{1}{c}{\begin{tabular}[c]{@{}c@{}}82.3/50\\ \textbf{84}/50\end{tabular}} \\
 \multirow{5}{*}{ResNet-18} & C-10/Gaussian & \begin{tabular}[c]{@{}c@{}}MSP\\ VoG\end{tabular} & \multicolumn{1}{c}{\begin{tabular}[c]{@{}c@{}}\textbf{92.7}/50\\ 85.1/50\end{tabular}} & \multicolumn{1}{c}{\begin{tabular}[c]{@{}c@{}}\textbf{95.1}/50\\ 90.6/50\end{tabular}} & \multicolumn{1}{c}{\begin{tabular}[c]{@{}c@{}}\textbf{88.2}/50\\ 73/50\end{tabular}} \\
 & C-10/iSUN & \begin{tabular}[c]{@{}c@{}}MSP\\ VoG\end{tabular} & \multicolumn{1}{c}{\begin{tabular}[c]{@{}c@{}}85.5/50\\ \textbf{92.3}/50\end{tabular}} & \multicolumn{1}{c}{\begin{tabular}[c]{@{}c@{}}89/52.8\\ \textbf{94.2}/52.8\end{tabular}} & \multicolumn{1}{c}{\begin{tabular}[c]{@{}c@{}}79.9/47.2\\ \textbf{89.3}/47.2\end{tabular}} \\
 & C-10/Tiny-ImageNet-Resize & \begin{tabular}[c]{@{}c@{}}MSP\\ VoG\end{tabular} & \multicolumn{1}{c}{\begin{tabular}[c]{@{}c@{}}84.7/50\\ \textbf{91.6}/50\end{tabular}} & \multicolumn{1}{c}{\begin{tabular}[c]{@{}c@{}}87.4/50.5\\ \textbf{93.1}/50.5\end{tabular}} & \multicolumn{1}{c}{\begin{tabular}[c]{@{}c@{}}79.8/49.5\\ \textbf{89.5}/49.5\end{tabular}} \\
 & C-10/LSUN-Resize & \begin{tabular}[c]{@{}c@{}}MSP\\ VoG\end{tabular} & \multicolumn{1}{c}{\begin{tabular}[c]{@{}c@{}}84.3/50\\ \textbf{92.3}/50\end{tabular}} & \multicolumn{1}{c}{\begin{tabular}[c]{@{}c@{}}86.4/50\\ \textbf{93.6}/50\end{tabular}} & \multicolumn{1}{c}{\begin{tabular}[c]{@{}c@{}}80/50\\ \textbf{90.4}/50\end{tabular}} \\\hline
 & C-10/C-100 & \begin{tabular}[c]{@{}c@{}}MSP\\ VoG\end{tabular} & \multicolumn{1}{c}{\begin{tabular}[c]{@{}c@{}}91.4/50\\ \textbf{93.1}/50\end{tabular}} & \multicolumn{1}{c}{\begin{tabular}[c]{@{}c@{}}93.1/50\\ \textbf{94.3}/50\end{tabular}} & \multicolumn{1}{c}{\begin{tabular}[c]{@{}c@{}}88.5/50\\ \textbf{91}/50\end{tabular}} \\
 \multirow{5}{*}{DenseNet-BC} & C-10/Gaussian & \begin{tabular}[c]{@{}c@{}}MSP\\ VoG\end{tabular} & \multicolumn{1}{c}{\begin{tabular}[c]{@{}c@{}}\textbf{95.8}/50\\ 88.2/50\end{tabular}} & \multicolumn{1}{c}{\begin{tabular}[c]{@{}c@{}}\textbf{97.3}/50\\ 93.4/50\end{tabular}} & \multicolumn{1}{c}{\begin{tabular}[c]{@{}c@{}}\textbf{92.7}/50\\ 74.3/50\end{tabular}} \\
 & C-10/iSUN & \begin{tabular}[c]{@{}c@{}}MSP\\ VoG\end{tabular} & \multicolumn{1}{c}{\begin{tabular}[c]{@{}c@{}}\textbf{92.8}/50\\ 92.5/50\end{tabular}} & \multicolumn{1}{c}{\begin{tabular}[c]{@{}c@{}}\textbf{95}/52.8\\ 94.9/52.8\end{tabular}}  & \multicolumn{1}{c}{\begin{tabular}[c]{@{}c@{}}\textbf{88.9}/47.2\\ 86.5/47.2\end{tabular}}  \\
 & C-10/Tiny-ImageNet-Resize & \begin{tabular}[c]{@{}c@{}}MSP\\ VoG\end{tabular} & \multicolumn{1}{c}{\begin{tabular}[c]{@{}c@{}}\textbf{91.3}/50\\ 90.6/50\end{tabular}}  & \multicolumn{1}{c}{\begin{tabular}[c]{@{}c@{}}\textbf{93.1}/50.5\\ 92.6/50.5\end{tabular}} & \multicolumn{1}{c}{\begin{tabular}[c]{@{}c@{}}\textbf{88.2}/49.5\\ 86.1/49.5\end{tabular}} \\
 & C-10/LSUN-Resize & \begin{tabular}[c]{@{}c@{}}MSP\\ VoG\end{tabular} & \multicolumn{1}{c}{\begin{tabular}[c]{@{}c@{}}92.9/50\\ \textbf{93}/50\end{tabular}} & \multicolumn{1}{c}{\begin{tabular}[c]{@{}c@{}}94.7/50\\ \textbf{94.9}/50\end{tabular}} & \multicolumn{1}{c}{\begin{tabular}[c]{@{}c@{}}\textbf{90}/50\\ 88.2/50 \end{tabular}}\\\hline
\end{tabular}
\label{tab:baseline}
\end{sc}
\end{table*}


\end{document}